\documentclass[sigconf]{acmart}
\usepackage[utf8]{inputenc}
\pdfoutput=1
\usepackage{soul}
\usepackage{url}
\usepackage{lipsum}
\usepackage{float}
\usepackage{amsmath}
\usepackage{amsthm}
\usepackage{booktabs}
\usepackage{algorithm}
\usepackage[noend]{algorithmic}
\usepackage{subcaption}
\usepackage{comment}
\usepackage{xspace}
\usepackage{placeins}
\usepackage{chngcntr}
\usepackage{balance}
\usepackage{enumitem,hyperref}
\usepackage{thmtools, thm-restate}

\theoremstyle{definition}

\newtheorem{defn}{\textbf{Definition}}
\newtheorem{prob}{\textbf{Problem}}

\newtheorem{thm}{\textbf{Theorem}}

\newtheorem{cor}{\textbf{Corollary}}
 
\setlist{nolistsep,leftmargin=*}
\newcommand{\rns}{\ensuremath\overrightarrow\CG\xspace} 
\newcommand{\mbd}[1]{\mathbf{#1}\xspace}
\newcommand{\z}{\mathbf{z}\xspace}

\newcommand{\name}{\textsc{Frigate}\xspace}

\newcommand{\gnn}{\textsc{Gnn}\xspace}
\newcommand{\mlp}{\textsc{Mlp}\xspace}

\newcommand{\gnns}{\textsc{Gnn}s\xspace}

\newcommand{\lstm}{\textsc{Lstm}\xspace}
\newcommand{\lstms}{\textsc{Lstm}s\xspace}

\newcommand{\gin}{\textsc{Gin}\xspace}

\newcommand{\mydefine}{\ensuremath\stackrel{\mathbf{\varDelta}}{=}\xspace}

\newcommand{\m}{\mathbf{m}\xspace}

\newcommand{\h}{\mathbf{h}\xspace}
\newcommand{\ii}{\mathbf{i}\xspace}
\newcommand{\f}{\mathbf{f}\xspace}
\newcommand{\g}{\mathbf{g}\xspace}
\newcommand{\oo}{\mathbf{o}\xspace}
\newcommand{\bc}{\mathbf{c}\xspace}
\newcommand{\W}{\mathbf{W}\xspace}
\newcommand{\w}{\mathbf{w}\xspace}
\newcommand{\oplstm}{\ensuremath\operatorname{LSTM}\xspace}

\newcommand{\x}{\mathbf{x}\xspace}

\newcommand{\s}{\mathbf{s}\xspace}
\newcommand{\q}{\mathbf{q}\xspace}
\newcommand{\y}{\hat{y}\xspace}

\newcommand{\bb}{\mathbf{b}\xspace}

\newcommand{\CG}{\mathcal{G}\xspace}

\newcommand{\CN}{\mathcal{N}\xspace}

\newcommand{\CV}{\mathcal{V}\xspace}
\newcommand{\CE}{\mathcal{E}\xspace}

\newcommand{\CL}{\mathbf{L}\xspace}

\newcommand{\VG}{\overrightarrow{\CG}}
\usepackage{array}
\newcolumntype{L}[1]{>{\raggedright\arraybackslash}m{#1}}

\newcolumntype{R}[1]{>{\raggedleft\arraybackslash}m{#1}}

\newcolumntype{C}[1]{>{\centering\arraybackslash}m{#1}}
\newcommand{\rev}[1]{\begingroup\color{black}#1\endgroup}

\DeclareMathOperator*{\minimize}{minimize}

\date{\today}
\newcommand{\cross}{\(\times\)}
\newcommand{\checkm}{\(\checkmark\)}
\title{\name: Frugal Spatio-temporal Forecasting on Road Networks}
\author{Mridul Gupta}
\affiliation{%
    \department{Yardi School of AI}
    \institution{Indian Institute of Technology, Delhi}
    \country{India}}
    \email{Mridul.Gupta@scai.iitd.ac.in}
\author{Hariprasad Kodamana}
\affiliation{%
\department{Yardi School of AI}
    \institution{Indian Institute of Technology, Delhi}
    \country{India}}
    \email{kodamana@iitd.ac.in} 
\author{Sayan Ranu}
\affiliation{%
\department{Yardi School of AI}
    \institution{Indian Institute of Technology, Delhi}
    \country{India}}
    \email{sayanranu@cse.iitd.ac.in}
\date{January 2023}
\copyrightyear{2023}
\acmYear{2023}
\setcopyright{acmlicensed}\acmConference[KDD '23]{Proceedings of the 29th ACM SIGKDD Conference on Knowledge Discovery and Data Mining}{August 6--10, 2023}{Long Beach, CA, USA}
\acmBooktitle{Proceedings of the 29th ACM SIGKDD Conference on Knowledge Discovery and Data Mining (KDD '23), August 6--10, 2023, Long Beach, CA, USA}
\acmPrice{15.00}
\acmDOI{10.1145/3580305.3599357}
\acmISBN{979-8-4007-0103-0/23/08}                                             
\begin{abstract}
    Modelling spatio-temporal processes on road networks is a task of growing importance. 
    While significant progress has been made on developing spatio-temporal graph neural networks (\gnns), existing works are built upon three assumptions that are not practical on real-world road networks. First, they assume sensing on every node of a road network. In reality, due to budget-constraints or sensor failures, all locations (nodes) may not be equipped with sensors. Second, they assume that sensing history is available at all installed sensors. This is unrealistic as well due to sensor failures, loss of packets during communication, etc. Finally, there is an assumption of static road networks. Connectivity within networks change due to road closures, constructions of new roads, etc. In this work, we develop \name to address all these shortcomings. \name is powered by a spatio-temporal \gnn that integrates positional, topological, and temporal information into rich \textit{inductive} node representations. The joint fusion of this diverse information is made feasible through a novel combination of \textit{gated} \textit{Lipschitz} embeddings with \lstms. We prove that the proposed \gnn architecture is provably more expressive than message-passing \gnns used in state-of-the-art algorithms. The higher expressivity of \name naturally translates to superior empirical performance conducted on real-world network-constrained traffic data. In addition, \name is robust to frugal sensor deployment, changes in road network connectivity, and temporal irregularity in sensing.
\end{abstract}
\begin{document}

\maketitle
\section{Introduction and Related Work}
\begin{figure}[t]
    \centering
        \includegraphics[width=3.4in]{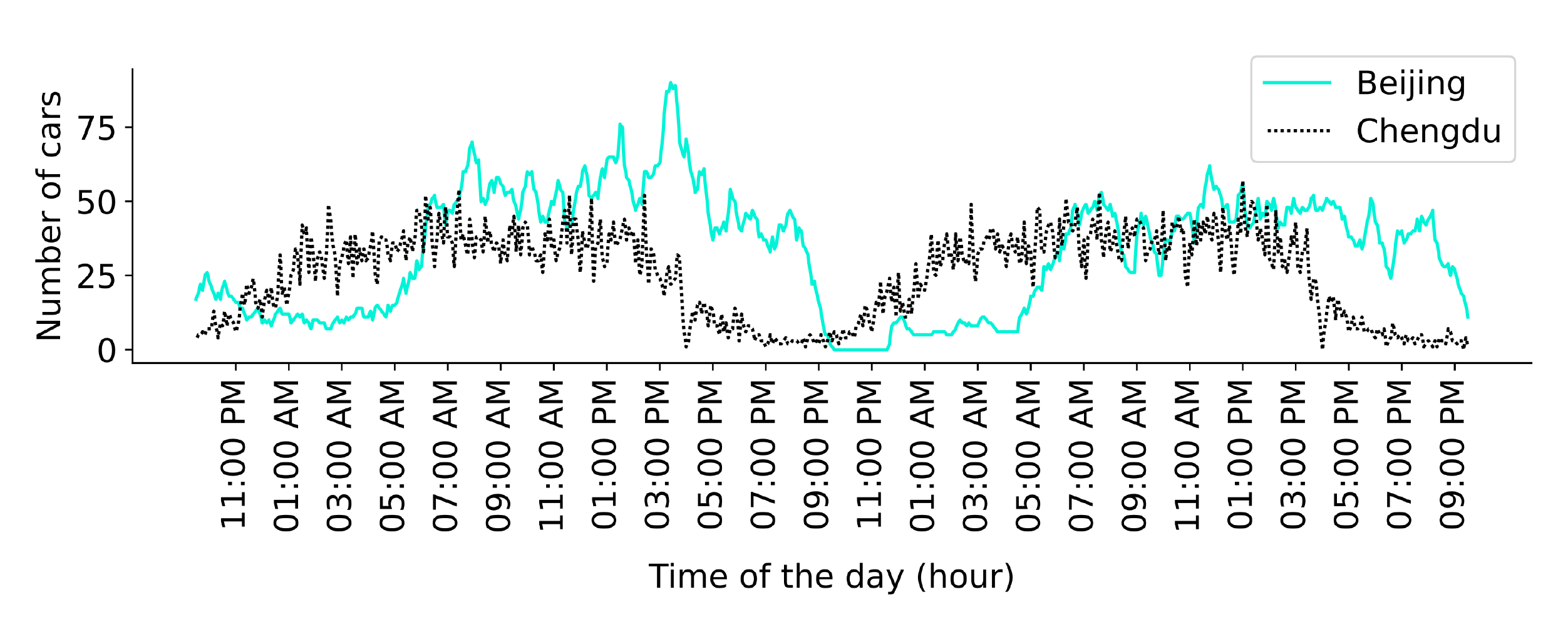}
        \vspace{-0.30in}
\caption{\label{fig:nodetimeseries}Snapshots of selected roads from Beijing and Chengdu displaying the tidal variation of traffic}
        \vspace{-0.20in}
\end{figure}
A road network can be modeled as a graph where nodes indicate important intersections in a region and edges correspond to streets connecting two intersections~\cite{skygraph,skyroute,netclus,tops,foodmatch,medya2018noticeable}. 
Modeling the evolution of spatio-temporal events on road networks has witnessed significant interest in the recent past~\cite{dcrnn,stgcn,stgode, stnn,wavenet}. 
Specifically, each node (or edge) participates in a time-series. This time-series is a function of not only the time of the day, but also the events unfolding in other nodes of the network. The objective of the forecasting task is to model the time-series evolution in each node and predict the values in the immediate future, for e.g., the next one hour. As examples, in Fig.~\ref{fig:nodetimeseries}, we present the number of cars passing through two randomly selected intersections in the cities of Beijing and Chengdu. As can be seen, there is significant variation in traffic through out the day. Furthermore, the time-series vary across nodes making the forecasting problem non-trivial.
\looseness=-1

One may learn an auto-regressive model independently at each node to fit the time-series data. This strategy, however, is limited by two critical factors. First, this ignores the connectivity-induced time-series dependency among nodes. For example, if one particular intersection (node) is observing traffic congestion, it is likely for neighboring nodes connected through an outgoing edge to be affected as well. Second, the number of parameters grows linearly with the number of nodes in the graph, making it non-scalable. 

\subsection{Existing work}
\label{sec:existingwork}
To address these specific needs of modeling network-dependent spatio-temporal processes, several algorithms merging models for structural data such as \gnns~\cite{stgcn,dstagnn,stfgnn} or Convolutional neural networks~\cite{stnn} along with auto-regressive architectures have been developed. The proposed work is motivated based on some assumptions made by existing algorithms that are not realistic for real-world road networks. Table~\ref{tbl:baselines} summarizes these. 
\begin{itemize}
\item \textbf{Ability to extrapolate from partial sensing:} Several of the existing algorithms assume that a sensor is placed in each node of the road network~\rev{\cite{agcrn,gman,wavenet,dstagnn,stsgcn,zigzag,stgode,stfgnn,zhou2023towards,gmsdr,mtgnn}}. Forecasting is feasible \textit{only} on these nodes. In reality, deploying and maintaining sensors across all intersections may be prohibitively expensive and cumbersome. Hence, it is important to also forecast accurately on nodes that do not have an explicit sensor placed on them. In this work, we show that this is indeed feasible by exploiting the network-induced dependency between nodes.
\item \textbf{Ability to absorb network updates:} The connectivity within a road network may change with time. The directionality of edges may change based on time of the day, new roads may get constructed adding new nodes and edges to the network, and existing road may get removed (temporarily or permanently) to accommodate emergent needs such as street festivals,  construction activities, flooding, etc. Under these circumstances, it is important to absorb small changes in the network topology without the need to retrain from scratch. Several models~\rev{\cite{agcrn,gman,wavenet,dstagnn,stsgcn,zigzag,wang2018real,zhou2023towards,gmsdr,mtgnn}} fail to absorb any updates since they are \textit{transductive} in nature, i.e., the number of parameters in their model is a function of the network size (number of nodes or edges). In this work, we develop an inductive model, which decouples the model parameter size from that of the network. Hence, accommodating changes to network structure does not require re-training.
\item \textbf{Ability to predict without temporal history or regularity:} Several of the existing algorithms can forecast on the time-series of a node \textit{only} if the data in the past $x$ time instants are available~\cite{stgode,stfgnn}. This limitation often arises from a methodology where dependency between all nodes is learned by computing similarity between their past time-series information. If past data is not available, then this similarity cannot be computed. Furthermore, some algorithms model the spatio-temporal road network as a 3-dimensional tensor, under the implicit assumption that temporal data is collected at a regular granularity (such as every five minutes)~\cite{stnn,stgode}. In reality, sensors may fail and may therefore provide data at irregular intervals or having no temporal history in the past $x$ time instants. For a model to be deployable in real workloads, it is pertinent to be robust to sensor failures.   
\end{itemize}
\begin{table}[t]
    \centering    \caption{\label{tbl:baselines}\rev{Baselines}}
    {\scriptsize
    \begin{tabular}{lp{1cm}p{1.6cm}p{2cm}}
        \toprule
        Algorithm  & Partial sensing & Absorb network updates &  Predict without temporal history \\
        \midrule
        AGCRN~\cite{agcrn} & \cross & \cross & \cross \\ 
        STGODE~\cite{stgode} & \cross & \cross & \cross \\ 
        DSTAGNN~\cite{dstagnn} & \cross & \cross & \cross \\ 
        STFGNN~\cite{stfgnn} & \cross & \cross & \cross \\ 
        \rev{G2S~\cite{zhou2023towards}} & $\rev{\times}$ & $\rev{\times}$ & $\rev{\times}$ \\
        \rev{GMSDR~\cite{gmsdr}} & $\rev{\times}$ & $\rev{\times}$ & $\rev{\times}$\\
        Z-GCNETs~\cite{zigzag} & \cross & \cross & \checkm \\
        STSGCN~\cite{stsgcn} & \cross & \cross & \checkm \\ 
        GraphWavenet~\cite{wavenet} & \cross & \cross & \checkm \\ 
        GMAN~\cite{gman} & \cross & \cross & \checkm \\ 
        \rev{MTGNN~\cite{mtgnn}} & $\rev{\times}$ & $\rev{\times}$ & $\rev{\checkmark}$ \\ 
        \rev{TISV~\cite{wang2018real}} & $\rev{\checkmark}$ & $\rev{\times}$ & $\rev{\checkmark}$ \\
        STGCN~\cite{stgcn} & \checkm & \checkm & \checkm \\ 
        DCRNN~\cite{dcrnn} & \checkm & \checkm & \checkm \\
        STNN~\cite{stnn} & \checkm & \checkm & \checkm \\
        \rev{LocaleGN~\cite{li2022few}} & $\rev{\checkmark}$ & $\rev{\checkmark}$ & $\rev{\checkmark}$\\
        \rev{ST-GAN~\cite{wang2022inferring}} & $\rev{\checkmark}$ & $\rev{\checkmark}$ & $\rev{\checkmark}$\\ 
        \bottomrule
    \end{tabular}}
\end{table}
\vspace{-0.10in}
\subsection{Contributions}
\label{sec:contributions}
Motivated by the above limitations, in this work, we ask the following questions: \textit{Is it possible to design an accurate and inductive forecasting model across all nodes in a graph based on partial sensing through a small subset of nodes? In addition, can the model predict on nodes with irregular time-series visibility, or in the worst case, no visibility at all?} We show that this is indeed possible through \name: \underline{FR}ugal and \underline{I}nductive Lipschitz-\underline{GA}ted Spatio-\underline{TE}mporal \gnn. Specifically, we make the following contributions:
\begin{itemize}
    \item \textbf{Problem formulation:} We formulate the problem of time-series forecasting on road networks. Taking a deviation from current works, the proposed formulation is cognizant of practical challenges such as limited sensor access, sensor failures, and noise in data (\S~\ref{sec:formulation}).
    \looseness=-1
    \item {\bf Novel architecture:} To enable robust forecasting, we develop a novel spatio-temporal \gnn called \name. At a high-level, \name is a joint architecture composed of stacks of \textit{siamese} \gnns and \lstms in encoder-decoder format (\S~\ref{sec:frigate}). Under the hood, \name incorporates several innovations. First, to succeed in accurate forecasting under frugal sensor deployment, \name uses \textit{Lipschitz-gated} attention over messages. Gating allows \name to receive messages from far-away neighbors without over-smoothing the node neighborhoods. In addition, Lipschitz embedding allows \name to inductively embed positional information in node representations. Second, \name is inductive and hence does not require re-training on network updates. Finally, the coupling between the \gnn and \lstm is devoid of any assumption on the temporal granularity. Hence, it is robust to sensor failures or irregular data collection. 
    \item {\bf Empirical evaluation:} We perform extensive evaluation on real-world road network datasets from three large cities. The proposed evaluation clearly demonstrates the superiority of \name over state-of-the-art algorithms and establishes its robustness to data noise and network changes (\S~\ref{sec:experiments}).
\end{itemize}
\section{Problem Formulation}
\label{sec:formulation}
In this section, we define the concepts central to our work and formulate our problem. All key notations used in our work are summarized in Table~\ref{tab:notation} in Appendix.
\vspace{-0.05in}
\begin{defn}[Road Network Snapshot]
    \label{def:graph_snapshot}
    \textit{A road network snapshot is represented as a directed graph $\CG = (\CV_t, \CE_t, \delta, \tau_t)$, where $\CV_t$ is the set of nodes representing road intersections at time $t$, $\CE_t \subseteq \CV_t \times \CV_t$ is the set of edges representing road segments at time $t$, a distance function $\delta : \CE_t \rightarrow \mathbb{R}$ representing the length (weight) of each road segment (edge), and the sensor readings $\tau_t =\left\{\tau^v_t\in \mathbb{R} \mid v\in\CV_t\right\}$ for each node at time $t \in \mathbb{N}$.}
\end{defn}
\vspace{-0.05in}

Intuitively, a road network snapshot characterizes the state of the road network at time $t$. We use $\tau_t^v$ to denote the sensor reading at node $v$. When $\tau_t^v$ is not available, either due to non-availability of a sensor at node $v$, or due to sensor failure, we assume $\tau_t^v$ is marked with a special label. We use $\tau_t^v=\varnothing$ to denote a missing value. Furthermore, in the real world, the sensor value $\tau_t^v$ may not be recorded exactly at time $t$. We thus assume $\tau_t^v$ to be the latest value since the last snapshot; $\tau_t^v=\varnothing$ if nothing has been recorded since the last snapshot. Note that we also make the vertex and edge sets time-dependent to account for the fact that minor changes are possible on the topology over time. Examples include changing directionality of traffic on a particular road (edge), addition of new roads and intersections, temporary road closures due to street festivals, etc. We use the notation $e=(u,v)$ to denote a road segment (edge) from node $u$ to $v$ and its length is denoted by $\delta(e)$. The length $\delta(e)$ of an edge $e$ is the \emph{Haversine} distance from the locations represented by $u$ and $v$.
\vspace{-0.05in}
\begin{defn}[Road Network Stream]
\label{def:graph}
\textit{A road network stream is the chronologically ordered set of snapshots of the road network taken at various time instances, $\VG = \{\CG_{1},\CG_{2},\dotsc, \CG_{T}\}$, where 
\(\CG_{t}\) is the road network snapshot at time instant $t$.}
\end{defn}
\vspace{-0.05in}
    
\begin{figure*}
    \centering
    \includegraphics[width=.9\linewidth]{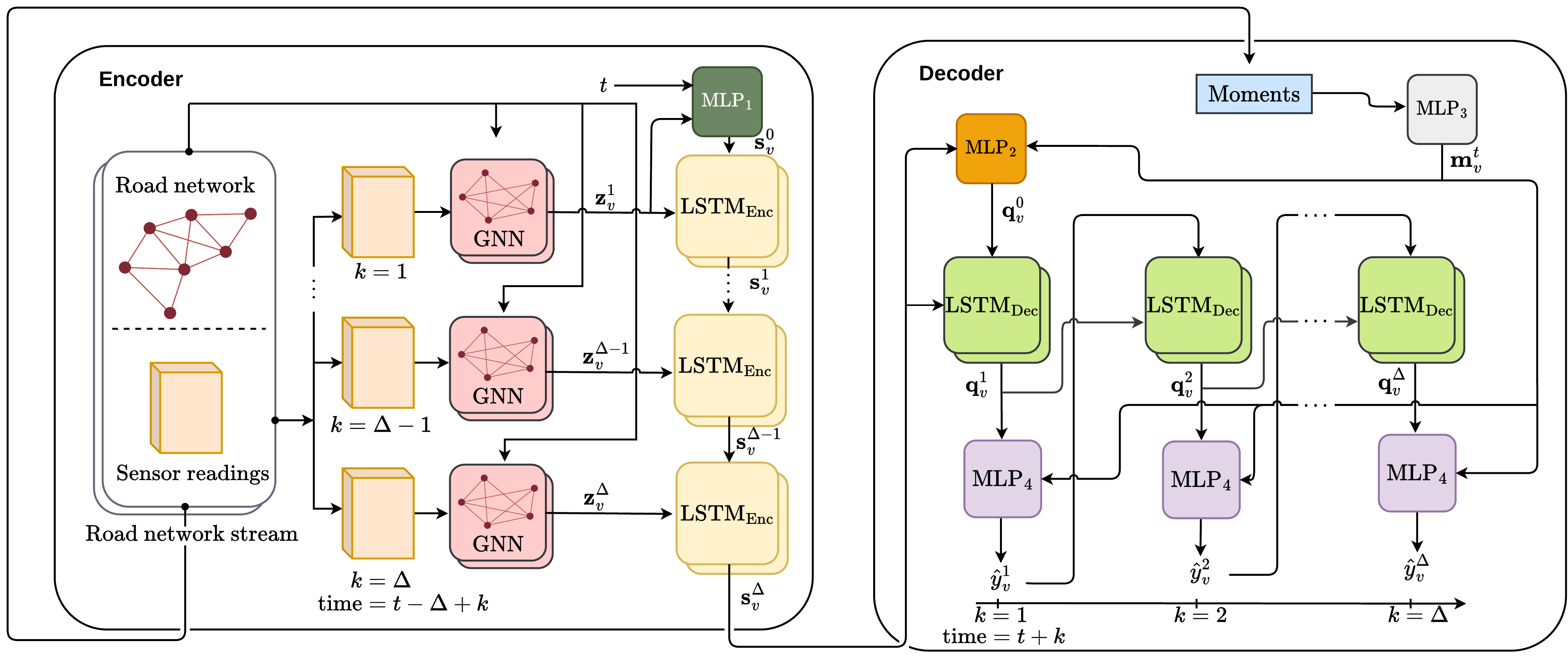}
    \vspace{-0.10in}
    \caption{\label{fig:model}The architecture of \name. Each of the neural blocks that are in same color are in siamese in nature. }
\end{figure*}

We assume, that $\forall t,\;\CV_{t}\approx\CV_{t+1}$ and similarly $\CE_{t}\approx\CE_{t+1}$. This assumptions are realistic based on real-world knowledge that change events on roads are rare over both space and time.

Our goal is to learn the dynamics of the time-series at each node and forecast future values. Towards that, the modeling problem is defined as follows.
\vspace{-0.05in}
\begin{prob}[Forecasting on Road Network]\hfill
\label{prob:frigate}

\noindent
\textbf{Training:} \textit{Given a road network stream $\VG = \left\{\CG_{1},\dotsc, \CG_{T}\right\}$, a forecasting horizon $\Delta$, and a timestamp $t$ where $\Delta\leq t\leq T-\Delta$, learn a function $\Psi$, parameterized by $\Theta$, to minimize the mean absolute prediction error over sensor values. Specifically,
\vspace{-0.05in}
\begin{equation}
\label{eq:mae}
{\small
\displaystyle{\minimize\left\{\frac{1}{|\CV^{tr}| \Delta }\sum_{k=1}^{\Delta} \sum_{v\in \CV^{tr}_{t+k}} \left\lvert\Psi_{\Theta}\left(v,t+k\:|\:\overrightarrow\CG_{[1,t]}\right)- \tau^v_{t+k}\right\rvert\right\}}
}
\end{equation}
Here, $\CV^{tr}_{t+k}=\left\{v\in\CV_{t+k}\mid\tau^v_{t+k}\neq\varnothing\right\}$ is the set of training nodes with ground truth sensor values at time $t+k$, $\overrightarrow\CG_{[1,t]}=\{\CG_1,\cdots,\CG_{t}\}$ denotes the subset of snapshots till $t$, and $\Psi_{\Theta}\left(v,t+k\:|\:\overrightarrow\CG_{[1,t]}\right)$ predicts the sensor value for node $v$ at time $t+k$ when conditioned on all snapshots till now, i.e., $\overrightarrow\CG_{[1,t]}$. We assume that to predict on a time horizon of $\Delta$ snapshots, we must train on a history of at least $\Delta$ snapshots.}

\noindent
\textbf{Inference:} \textit{Let the last recorded snapshot be at time $t$. Given node $v\in\CV_{t}$ and forecasting horizon $\Delta$, compute:} 
\vspace{-0.05in}
\begin{equation}
\forall k\in [1,\Delta],\;\left\{\Psi_{\Theta}\left(v,t+k\:|\:\VG_{[1,t]}\right)\right\}
\end{equation}
\end{prob}
\noindent
In addition to predicting accurately, $\Psi_{\Theta}$, must also satisfy the following properties:
\looseness=-1
\begin{itemize}
\item \textbf{Inductivity:} The number of parameters in model $\Psi_{\Theta}$, denoted as $\lvert \Theta \rvert$, should be independent of the number of nodes in the road network at any given timestamp. Inductivity enables the ability to predict on unseen nodes without retraining from scratch. 
\item \textbf{Permutation invariance:} Given any \textit{permutation function} \\$\mathcal{P}\left(\VG_{[1,t]}\right)$ that randomly permutes the node set of each graph $\CG\in\VG_{[1,t]}$, we require: 
\vspace{-0.10in}
\begin{equation}
\nonumber
\forall k\in[1,\Delta],\;\Psi_{\Theta}\left(v,t+k\:|\:\VG_{[1,t]}\right)=\Psi_{\Theta}\left(v,t+k\:|\:\mathcal{P}\left(\VG_{[1,t]}\right)\right)
\end{equation}
More simply, if a graph contains $n$ nodes, then there are $n!$ possible permutations over its node set, and hence that many adjacency matrix representations of $\CG_t$. Permutation invariance ensures that the output is dependent only on the topology of the graph and not coupled to one of its specific adjacency matrix representations. Hence, it aids in generalizability while also ensuring that the location of change in the adjacency matrix owing to a node/edge insertion or deletion is inconsequential to the output.
\looseness=-1

\item {\bf Semi-supervised learning: } 
Towards our objective of frugal forecasting,  
$\Psi_{\Theta}(v,t\mid \VG_{[1,t]})$ should be computable even if temporal features from past snapshots are not available on $v$ as long as \textit{some} nodes in the graph contain temporal information.
\end{itemize}

\noindent
\section{\name: Proposed Methodology}
\label{sec:frigate}
At an individual node level, the proposed problem is a \textit{sequence-to-sequence} regression task, wherein we feed the time-series over past snapshots and forecast the future time-series on the target node. To model this problem, we use an \textit{Encoder-Decoder} architecture as outlined in Fig.~\ref{fig:model}. To jointly capture the spatio-temporal dependency among nodes, the encoder is composed of a stack of \textit{siamese} \gnns and \lstms, i.e., the weights and architecture are identical across each stack. The number of stacks corresponds to the number of historical snapshots one wishes to train on. Each stack is assigned an index depending on how far back it is from the current time $t$. Due to the siamese architecture, the number of stacks does not affect the number of parameters. In addition, the siamese design allows one to dynamically specify the stack size at inference time in a query-specific manner depending on the amount of data available. For simplicity, we will assume the number of historical snapshots to be the same as the forecasting horizon, which is denoted as $\Delta$.

In the $k^{th}$ stack, where $0\leq k\leq \Delta$, graph snapshot $\CG_{t-\Delta+k}$ is passed through the \gnn to learn node representations. The node representation not only characterises its own state, but also the state in its ``neighborhood''. The ``neighborhood'' that affects the time-series of the target node is automatically learned through \textit{Lipschitz-gating}. Each stack of siamese \lstm in the encoder receives two inputs: the node embedding of the target node corresponding to its timestamp and the \lstm output from the previous stack. 
\looseness=-1

The decoder has a stack of siamese \lstms as well. However, the decoder \lstms have separate weights than the encoder \lstms. The stack size is the same as the forecasting horizon $\Delta$. The output of a decoder \lstm is augmented with the \textit{moments}~\cite{moments} of the sensor value distribution in neighborhood of the target node, which injects a strong prior to the model and elevates its performance. Finally, this augmented representation is passed through an \mlp to predict the time-series value. The entire architecture is trained \textit{end-to-end} with \textit{mean absolute error (MAE)} loss function, as defined in Eq.~\ref{eq:mae}. The $\Psi_{\Theta}\left(v,t+k\:|\:\overrightarrow\CG_{[1,t]}\right)$ term in Eq.~\ref{eq:mae} is computed as:
\vspace{-0.05in}
\begin{equation}
\Psi_{\Theta}\left(v,t+k\:|\:\overrightarrow\CG_{[1,t]}\right)=\y_v^k
\end{equation}
where $k\in[1,\Delta]$ and $\y_v^k$, as depicted in Fig.~\ref{fig:model}, is the predicted output generated through the $k^{th}$ \lstm in the decoder. The next sections detail these individual components.
\looseness=-1 

%
%

%
%
%
\subsection{\gnn Module of \name}
\label{sec:gnn}
In this section, we discuss the architecture of a single stack of \gnn. We use an $L$-layered message-passing \gnn to model node dependencies. In each layer, each node $v$ receives messages from its neighbors. These messages are aggregated and passed through a neural network, typically an \mlp, to construct the representation of $v$ in the next layer. 

 The simplest aggregation framework would be a $\textsc{MeanPool}$ layer where the aggregated message at $v$ from its neighbors is simply the mean of their representations. However, this approach is too simplistic for the proposed problem. Specifically, a road network is directed, and hence drawing messages only through incoming (or outgoing) edges is not enough. On the other hand, a direction-agnostic message passing scheme by uniformly drawing messages through all incident edges fails to capture traffic semantics. Furthermore, not all edges are equally important. An incoming edge corresponding to an arterial road is likely to have much more impact on an intersection (node) than a road from a small residential neighborhood. Hence, the importance of a road (edge) towards an intersection (node) must be learned from the data, and accordingly, its message should be weighted. To capture these semantics, we formulate our message-passing scheme as follows.

 Let $\h^{\ell}_v$ denote the representation of node $v$ in layer $\ell\in [0,L]$ of the $k^{th}$ \gnn. Furthermore, the outgoing and incoming neighbors of $v$ are defined as $\CN^{out}_v=\{u\mid (v,u)\in\CE\}$ and $\CN^{in}_v=\{u\mid (u,v)\in\CE\}$ respectively. Now, $h^{\ell+1}_v$ is constructed as follows:
 \begin{align}
 \label{eq:hl}
 \h^{\ell+1}_v&= \sigma_1\left(\W^{\ell}_1 \left(h_v^{\ell}\parallel\m^{\ell,out}_v\parallel \m^{\ell,in}_v\right)\right) \text{, where}\\
 \m^{\ell,out}_v&=\textsc{Aggr}^{out}\left(\left\{\h_u^{\ell}\mid u\in N^{out}_v\right\}\right)\\
\m^{\ell,in}_v&=\textsc{Aggr}^{in}\left(\left\{\h_u^{\ell}\mid u\in N^{in}_v\right\}\right)
 \end{align}
$\parallel$ represents the \textit{concatenation} operation, $\W^{\ell}_1\in\mathbb{R}^{3d\times d}$ is a learnable weight matrix; $d$ is the dimension of node representations. More simply, we perform two separate aggregations over the messages received from the incoming and outgoing neighbors. The aggregated vectors are concatenated and then passed through a linear transformation followed by non-linearity through an activation function $\sigma_1$. In our implementation, $\sigma_1$ is \textsc{ReLU}. By performing two separate aggregations over the incoming and outgoing edges and then concatenating them, we capture both directionality as well as the topology. 
\looseness=-1

The aggregation functions perform a weighted summation, where the weight of a message-passing edge is learned through \textit{sigmoid gating} over its \textit{positional embedding}. More formally,

\vspace{-0.05in}
\begin{footnotesize}
\begin{align}
\label{eq:aggregation_out}
\textsc{Aggr}^{out}\left(\left\{\h_u^{\ell}\mid u\in N^{out}_v\right\}\right)&=\sum_{\forall u \in N^{out}_v} \beta_{v,u}\h^{\ell}_u\\
\label{eq:aggregation_in}
\textsc{Aggr}^{in}\left(\left\{\h_u^{\ell}\mid u\in N^{in}_v\right\}\right)&=\sum_{\forall u \in N^{in}_v} \beta_{u,v}\h^{\ell}_u\\
\label{eq:sigmoid}
\beta_{v_i,v_j}&=\frac{1}{1+e^{-\omega_{v_i,v_j}}}\text{, where}\\
\label{eq:edgescalar}
\omega_{v_i,v_j}&=\w^{\ell}\cdot\CL_{v_i,v_j}+b \text{, where}\\
\label{eq:edgeposition}
\CL_{v_i,v_j}&=\sigma_2\left(\W^{\ell}_{\CL}\left(\left(\w_{\delta}^T\cdot \delta(v_i,v_j)\right)\parallel \CL_{v_i}\parallel \CL_{v_j}\right)\right)%
\end{align}
\end{footnotesize}

Here, $\CL_v$ denotes the positional embedding of node $v$, whose construction we will discuss in Section~\ref{sec:lipschitz}. Intuitively, the positional embedding represents the location of a node such that if two nodes are within close proximity in terms of shortest path distance, then their positional embeddings should also be similar. Since, $\CL_{v_i,v_j}$ is a function of the positional embeddings of its two endpoints and the spatial distance between them, $\CL_{v_i,v_j}$ may be interpreted as the positional embedding of the edge $(v_i,v_j)$. The importance of an edge is computed by passing its positional representation through an \mlp (Eq.~\ref{eq:edgeposition}), which is subsequently converted to a scalar (Eq.~\ref{eq:edgescalar}). Finally, the scalar is passed through a sigmoid gate (Eq.~\ref{eq:sigmoid}) to obtain its weight. Here, $\w_{\delta}\in\mathbb{R}^{d_{\delta}}$, $\W^{\ell}_{\CL}\in\mathbb{R}^{2d_L+d_{\delta}\times d_{\CL_{e}}}$ and $\w^{\ell}\in\mathbb{R}^{d_{L_e}}$ are learnable weight parameters; $d_L$ and $d_{L_e}$ are the dimensionality of the positional embeddings of nodes and edges respectively, and $d_\delta$ is the dimension of the projection created over edge distance $\delta_{v_i,v_j}$.\footnote{We project the edge distance to a higher dimensional representation since otherwise the significantly larger number of dimensions allocated to positional embeddings of the endpoints may dominate the single dimension allocated for edge distance.}. $\sigma_2$ in Eq.~\ref{eq:edgeposition} is an activation function to apply non-linearity. $b$ in Eq~\ref{eq:edgescalar} is the learnable bias term.
\looseness=-1

\begin{figure}[t]
    \includegraphics[width=0.8\linewidth]{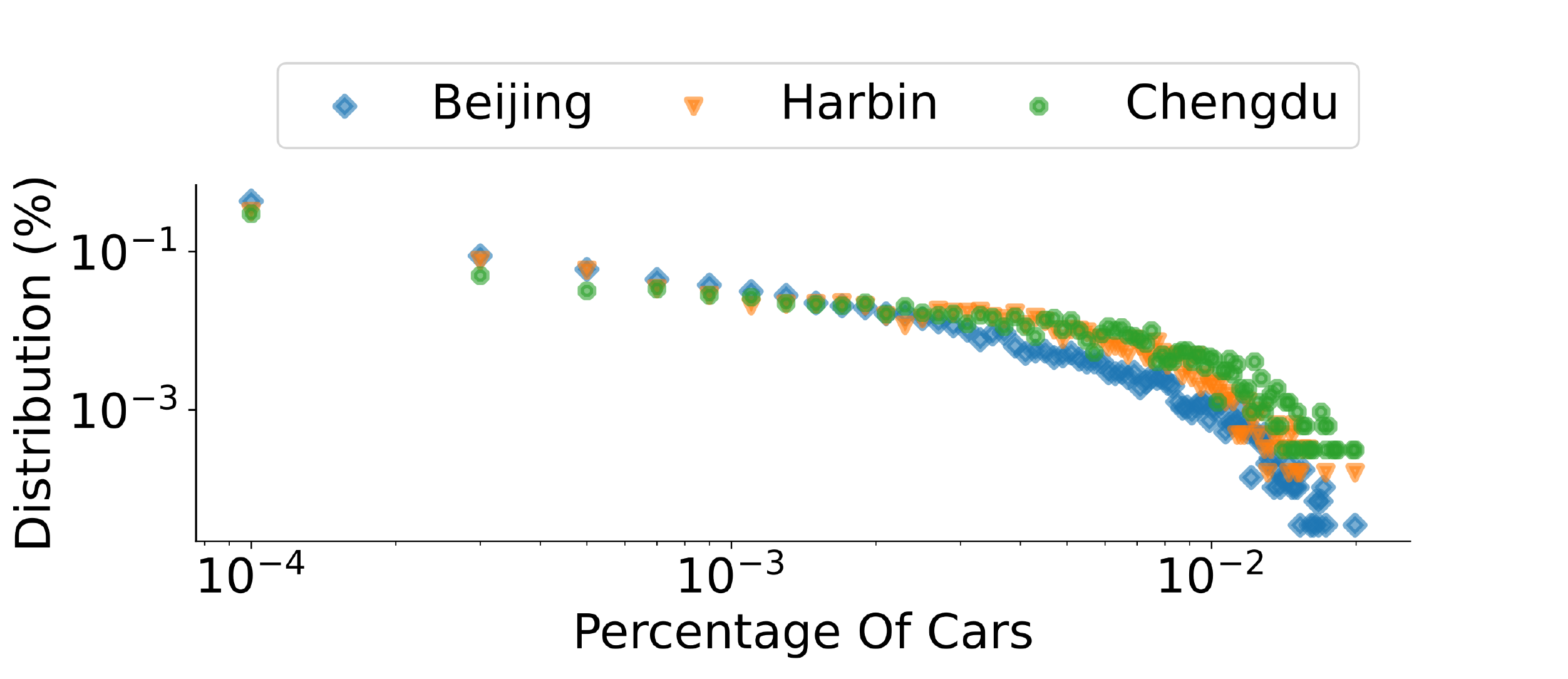}
    \vspace{-0.15in}
        \caption{\label{fig:traffic}The distribution of cars flowing through the edges in the cities of Beijing, Chengdu and Harbin.}
\end{figure}
We now discuss the rationale behind this design. In a road network, the distribution of traffic over edges resemble a \textit{power-law} (Fig.~\ref{fig:traffic}). While arterial roads carry a lot of traffic and therefore is a strong determinant of the state of the downstream roads, local roads, which forms the majority, have less impact. Hence, we need to learn this importance of roads from the data. Furthermore, we would like the \gnn to be deep without \textit{over-smoothing} or \textit{over-squashing} the representations~\cite{gnnbenchmark}. Sigmoid gating serves these dual requirements. First, it assigns a weight to every edge to determine its importance. Second, it enables us to go deep since as more layers are added, we receive messages only from the edges with high weights and therefore avoid over-squashing. Semantically, the roads with high weights should correspond to the arterial roads. Note that unlike Graph attention networks~\cite{gat}, where edges compete among each other for importance (due to normalization in \textsc{SoftMax}), we do not have that effect in a Sigmoid gating. This is more natural for road networks since the magnitude of representations on a busy intersection should be higher than an intersection with no incoming arterial roads. In addition, as we will formally prove in Section~\ref{sec:characterization}, Sigmoid gating provides higher expressive power. Finally, it is worth noting that the attention weights are directional since it is a function of the positional edge embedding where $\CL_{v_i,v_j}\neq \CL_{v_j,v_i}$.
\looseness=-1

\noindent
\textbf{Initial embedding of nodes:} Let $t$ be the current time and $\Delta$ the number of historical snapshots we are training on. Thus, we will have $\Delta$ stacks of \gnn 
 where the $k^{th}$ \gnn corresponds to snapshot of time $t-\Delta+k,\; k\in[0,\Delta]$ (Recall Fig.~\ref{fig:model}). The initial embedding of all nodes $v\in \CV_{t-\Delta+k}$ in the $k^{th}$ \gnn is defined as:
\begin{equation}
\label{eq:h0}
\h^{0,k}_v=\left(\w_{\tau}^T\cdot\tau^v_{t-\Delta+k} \right)\parallel\CL_v
\end{equation}
$\w_{\tau}$ is a learnable vector to map the sensor value to a higher dimensional space. $\CL_v$ is the positional embedding of $v$. Note we do not use the stack index in the notation for $\h^{\ell}_v$ (Eq.~\ref{eq:hl}) since all computations are identical except the time-dependent input in Eq.~\ref{eq:h0}. The final output after the $L^{th}$ layer in the $k^{th}$ \gnn is denoted as $\z^k_v=\h^{L,k}_v$.
\subsubsection{Positional Embeddings}
\label{sec:lipschitz}
The simplest approach to encode the position of a node is to embed its latitude and longitude into a higher-dimensional representation. This embedding, however, would not reflect the constraints imposed by the road network. Hence, to learn network-induced positional embeddings, we use \textit{Lipschitz embeddings}.
\begin{defn}[Lipschitz Embedding]
\label{def:lipschitz}
\textit{Let $\mathcal{A} = \{a_1,\cdots, a_m\} \subseteq \CV$ be a randomly selected subset of nodes. We call them anchors. Each node $v$ is embedded into an $m$-dimensional feature vector $\CL_v$ where $\CL_v[i]=\frac{sp(a_i,v)+sp(v,a_i)}{2}$, where $sp(u,v)$ is the shortest path distance from $u$ to $v$.} 
\end{defn}
\vspace{-0.05in}
\noindent
 The efficacy of the Lipschitz embedding lies in how well it preserves the shortest path distances among nodes. A well-accepted convention to measure the preservation of distances between the original space and the transformed space is the notion of \textit{distortion}~\cite{linial1995geometry}.
 \looseness=-1

\begin{defn}[Distortion]
\label{def:distortion}
    \textit{Let $X$ be a set of points embedded in a metric space with distance function $d_X$. Given a function \(f:X\rightarrow Y\) that embeds objects in $X$ from the finite metric space \((X,d_X)\) into another finite metric space \((Y,d_Y)\). We define:}
    \begin{align}
        \text{expansion}(f) &= \max_{x_1,x_2\in X} \frac{d_Y(f(x_1),f(x_1))}{d_X(x_1,x_2)}\\
        \text{contraction}(f) &= \max_{x_1,x_2\in X} \frac{d_X(x_1,x_2)}{d_Y(f(x_1),f(x_2))}
    \end{align}
    \textit{The \textit{distortion} of an embedding \(f\) is defined as the product of its expansion and contraction. If the distortion is $\alpha$, this means that $\forall\,x,y\in X,\;\frac{1}{\alpha}d_X(x,y)\le d_Y(f(x), f(y))\le d_X(x,y)$.}
\end{defn}
In the context of a road network, $X=\CV$, and $d_X(x_1,x_2)$ is the average two-way shortest path distance between $x_1$ and $x_2$ as defined in Def.~\ref{def:lipschitz}. Furthermore, the mapping $f:X\rightarrow Y$ is simply the Lipschitz embedding of $X$. Now, to define $d_Y(f(x_1),f(x_2))$, along with dimensionality $m$, we use \emph{Bourgain's Theorem}~\cite{linial1995geometry}.
\begin{thm}[Bourgain's Theorem~\cite{linial1995geometry}]
    \label{thm:bourgain}
    \textit{Given any finite metric space \((X, d_X)\), there exists a Lipschitz embedding of \((X, d_X)\) into \(\mathbb{R}^m\) under any \(L_p\) norm such that \(m=O(\log^2 n)\) and the distortion of the embedding is \(O(\log n)\), where \(n=\lvert X\rvert=|\CV|\).}
\end{thm}
Based on Bourgain's theorem, if we can show that $d_X(u,v)=\frac{sp(u,v)+sp(v,u)}{2}$, as defined in Def.~\ref{def:lipschitz}, is metric, then choosing \(m=O(\log^2 |\CV|)\) anchors and any $L_p$ distance in the embedded space would provide a distortion of \(O(\log |\CV|)\). We next show that $d_X(u,v)$ is indeed a metric.

\begin{restatable}{lem}{lemone}
\label{lem:metric_dist}
$d_X(u,v)=\frac{sp(u,v)+sp(v,u)}{2}$ \textit{is a metric.} 
\end{restatable}
\vspace{-0.05in}
\renewcommand{\qedsymbol}{}
 \begin{proof} \vskip-6pt\rev{\textit{See App.~\ref{app:proof_lem_one}.}}
 \end{proof}
 \looseness=-1
 
 The exact algorithm to choose the anchors is described in \cite{linial1995geometry}. Note that we use the same set of anchors across all snapshots. Since the topology may change across snapshots, the positional embeddings are time-varying as well. Although unlikely, it is possible for an anchor node to get deleted in a particular snapshot. In such a scenario, we denote the distance corresponding to this dimension as $\infty$ for all nodes. 

\subsection{Encoding Temporal Dependency}
To encode long-term temporal dependencies over the node representations, we use an \lstm encoder. Specifically, the final-layer outputs of the \gnn in the $k^{th}$ stack, denoted as $\z^k_v$, is fed to the $k^{th}$ \lstm stack. In addition, the $k^{th}$ \lstm also receives the hidden state of the $(k-1)^{th}$ \lstm (Recall Fig.~\ref{fig:model}). Mathematically, the output of the $k^{th}$ \lstm on node $v$, denoted as $\s^k_v\in\mathbb{R}^{d_{Lenc}}$, is computed as follows. 
\vspace{-0.10in}
\begin{equation}
\label{eq:lstm}
\s^{k}_v=
\begin{cases}
    \lstm_{Enc}\left(\s_v^{k-1},\z_v^{k}\right)&\text{if k>1}\\
    \lstm_{Enc}\left(\mlp_1\left(t,\z^k_v\right),\z_v^{k}\right) & \text{if k=1}
    \end{cases}
\end{equation}
Since $\s^{0}_v$ is undefined for the first \lstm, i.e., $k=1$, we make it a learnable vector through the $\mlp_1$. The output of the final \lstm stack corresponding to current time $t$ (See Fig.~\ref{fig:model}), feeds into the decoder. $d_{L_{enc}}$ is the dimension of the \lstm representations. 

Lines \ref{algl:enc_start}--\ref{algl:enc_end} in Alg.~\ref{alg:forwardpass} summarize the encoder component. Given stream $\VG$, current time $t$, and a target node $v$, we extract the subset $\VG_{[t-\Delta,t]}\subseteq\VG$ where $k$ denotes the number of \gnn-\lstm stacks. The processing starts at $\CG_{t-\Delta}$ where the $k^{th}$ \gnn stack embeds $v$ into $\z_v^k$. Next, $\z_v^k$ is passed to the $k^{th}$ \lstm. This completes one stack of computation. Iteratively, we move to the $(k+1)^{th}$ stack and the same operations are repeated till $k=\Delta$, after which $\s^{\Delta}_v$~(Eq.~\ref{eq:lstm}) is fed to the decoder.

\renewcommand{\algorithmicensure}{\textbf{Input:}}
\renewcommand{\algorithmicensure}{\textbf{Output:}} 
\begin{algorithm}[t]
    \caption{\name forward pass}
    \label{alg:forwardpass}
    {\scriptsize
    \begin{algorithmic}[1]
        \REQUIRE \name, stream $\rns$, target node $v$, current timestamp $t$, prediction horizon $\Delta$
        \ENSURE Predicted sensor values $\hat{y}^v_{t+1}\dotsc\hat{y}^v_{t+\Delta}$
        \STATE $\s_v^0\gets\mlp_1(t,\z_v^1)$\label{algl:enc_start}
        \STATE $k\gets 1$
        \FORALL{$\CG\in\rns_{[t-\Delta,t]}$}
            \STATE $\z_v^k\gets \gnn(v)$
            \STATE $\s_v^k \gets \oplstm_\text{Enc}(\s_v^{k-1},\z_v^{k})$
            \STATE $k\gets k+1$ \label{algl:enc_end}
        \ENDFOR
        \STATE $\m_v^t\gets\mlp_3\left(moments\left(\left\{\tau^{t'}_u \neq \varnothing| t'\in[t-\Delta,t], (u,v)\text{ or } (v,u)\in \CE^{t'}\right\}\right)\right)$ \label{algl:dec_begin}
        \STATE $\hat{y}_0\gets \mlp_2(\s_v^\Delta,\m_v^t)$
        \STATE $\q_v^0\gets\s_v^\Delta$
        \FORALL{$k\in[1 \dotsc \Delta]$}
            \STATE $\q_v^k\gets\oplstm_{\text{Dec}}(\q_v^{k-1},\hat{y}^{k-1}_v)$
            \STATE $\hat{y}_v^k\gets\mlp_4(\q_v^k,\m_v^t)$ \label{algl:dec_end}
        \ENDFOR
        \STATE Re-index $\hat{y}^v_k\mapsto\hat{y}^v_{t+k}$
        \RETURN $\hat{y}^v_{t+1}, \dots, \hat{y}^v_{t+\Delta}$
    \end{algorithmic}}
\end{algorithm}

\subsection{Decoder}
\label{sec:decoder}

The decoder is composed of a stack of $\Delta$ (forecasting horizon) siamese \lstms. Assuming $t$ to be the current time, the output of the $k^{th}$ \lstm in the decoder corresponds to the predicted sensor reading at time $t+k$. Each \lstm receives two inputs: (1) the predicted sensor value in the previous timestamp denoted as $\y_v^{k-1}\in\mathbb{R}$ and (2) the hidden state of the previous \lstm, denoted as $\q_v^{k-1}\in\mathbb{R}^{d_{dec}}$. Thus, the output of the $k^{th}$ \lstm is expressed as:
\vspace{-0.05in}
\begin{equation}
\label{eq:decoder}
\q^{k}_v=
\begin{cases}
    \lstm_{Dec}\left(\y_v^{k-1},\q_v^{k-1}\right)&\text{if k>1}\\
    \lstm_{Dec}\left(\mlp_2\left(\s^{\Delta}_v,\m_v^t\right),\s_v^\Delta\right) & \text{if k=1}
    \end{cases}
\end{equation}

We have a special case for $k=1$, since both $\y_v^{0}$ and $\q_v^{0}$ are undefined. Since we assume a setting of partial sensing across nodes, $\tau^t_v$ may not be available and hence cannot be used to substitute $\y_v^{0}$. To mitigate this situation, $\q_v^{k-1}$ is replaced with the output $\s^{\Delta}_v$ of the last \lstm in the encoder and $\y_v^{0}$ is estimated through the $\mlp_4$. This \mlp takes as input $\s^{\Delta}_v$ and a representation of the \textit{moments}~\cite{moments} of the observed sensor values in the neighborhood of $v$, denoted as $\m^t_v$. Formally,
\begin{align}
\nonumber
\resizebox{.97\linewidth}{!}{%
$\m^t_v=\mlp_3\left(moments\left(\left\{\tau^{t'}_u \neq \varnothing| t'\in[t-\Delta,t], (u,v)\text{ or } (v,u)\in \CE^{t'}\right\}\right)\right)$%
}
\end{align}
Finally, the predicted sensor value $\forall k\in [1,\Delta],\:\y_v^{k}$ is computed as:
\begin{equation}
\y_v^{k}=\mlp_4\left(\q_v^k,\m^t_v\right)
\end{equation}

Note that instead of directly predicting the sensor value from the \lstm hidden state $\q^k_v$, we augment this information with statistical information $\m^t_v$ on the time-series at $v$. This provides a strong inductive bias to the model and elevates its performance, which we will substantiate empirically during our ablation study in Section~\ref{sec:experiments}. Lines \ref{algl:dec_begin}--\ref{algl:dec_end} in Alg.~\ref{alg:forwardpass} summarize the computations in the decoder.
\subsection{Theoretical Characterization of \name}
\label{sec:characterization}

\textbf{Fact 1.} \textit{\name is inductive and permuation-invariant.}

\noindent 
\textsc{Discussion.} As outlined in Sections~\ref{sec:gnn}-\ref{sec:decoder} and summarized in Table~\ref{tab:parameters} in Appendix, the parameter-size is independent of the network size, number of snapshots being used for training, forecasting horizon, and the temporal granularity across snapshots. Furthermore, since \name uses sigmoid-weighted sum-pool, it is permutation invariant to node set. Consequently, \name can inherently adapt to changes in topology or forecast on unseen nodes without the need to re-train from scratch. As outlined in Table~\ref{tbl:baselines} and discussed in Sec.~\ref{sec:existingwork}, majority of existing works on spatio-temporal network-constrained forecasting do not satisfy the needs of being inductive and permutation-invariant. $\hfill\square$

We now focus on the expressive power of \name. Message-passing \gnns that aggregate messages from their neighbors are no more powerful than 1-WL~\cite{gin}. Hence, given nodes $v$ and $u$, if their $\ell$-hop neighborhoods are isomorphic, they will get identical embeddings. Thus, the position of a node plays no role. Methodologies such as DCRNN~\cite{dcrnn}, STGCN~\cite{stgcn}, or \rev{STGODE~\cite{stgode}}, that build on top of these message-passing \gnns will consequently inherit the same limitation. \name does not suffer from this limitation.

\begin{figure}
    \centering
    \includegraphics[width=0.4\linewidth]{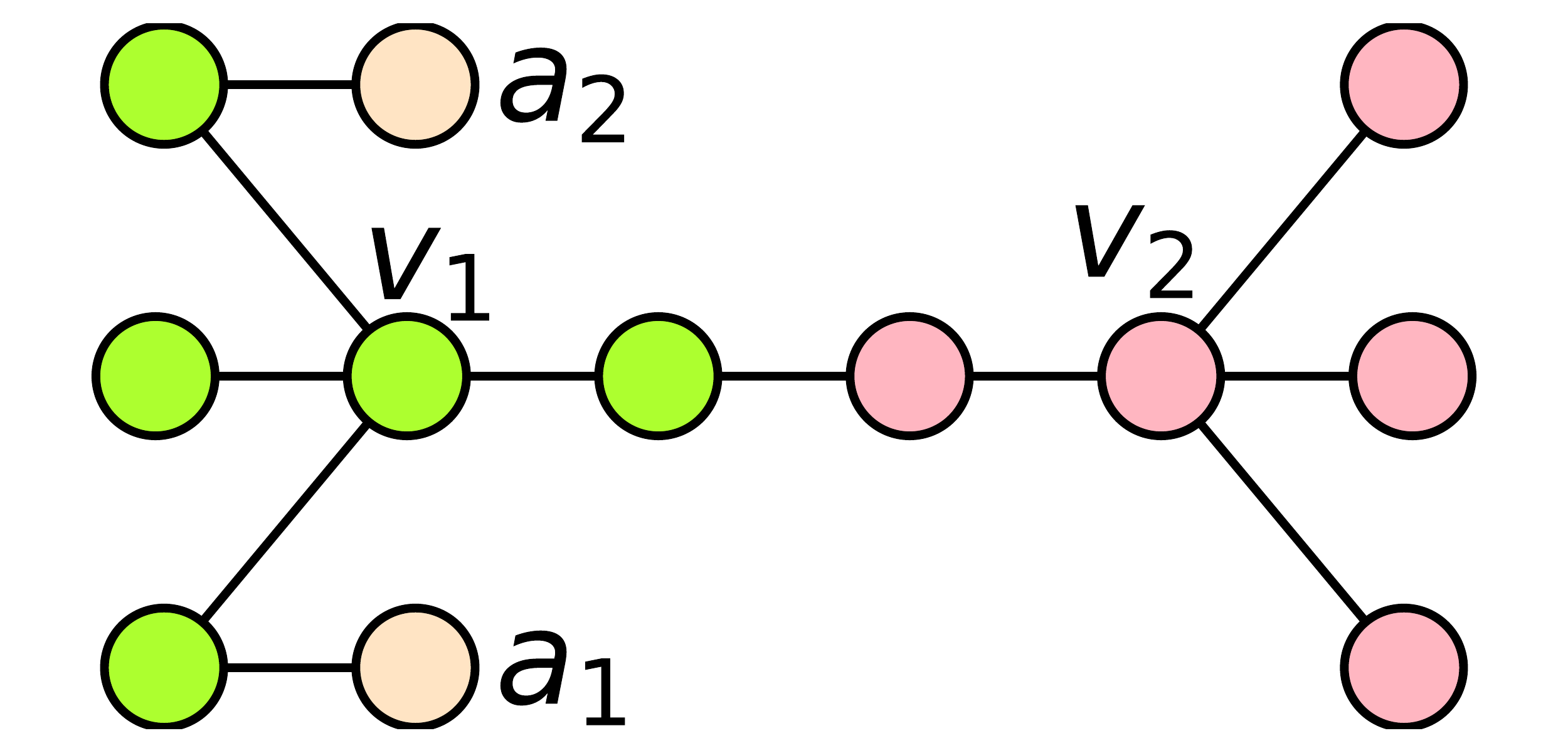}
    \caption{Sample graph to illustrate expressivity of \gnn in \name.}
    \label{fig:g3}
    \end{figure}

\renewcommand{\qedsymbol}{$\square$}
\begin{restatable}{lem}{lemtwo}
\label{lem:position}
\textit{\name can distinguish between isomorphic neighborhoods based on positioning.}
\end{restatable}
\begin{proof}
    Consider nodes $v_1$ and $v_2$ in Fig.~\ref{fig:g3}. If we use a 1-layered message-passing \gnn, then their embeddings would be identical since the 1-hop neighborhoods (color coded in green and pink) are isomorphic. Hence, DCRNN, STGCN, \rev{or STGODE} would not be able to distinguish between them. \name augments initial node features with Lipschitz embeddings. Assuming  $a_1$ and $a_2$ to be the anchor nodes, $v_1$ would have a Lipschitz embedding of $[2,2]$ vs. $[5,5]$ for $v_2$. Hence, \name would generate different embeddings for $v_1$ and $v_2$.
    \looseness=-1
\end{proof}
\renewcommand{\qedsymbol}{}

\vskip-6pt
As we will show in our ablation study, removing positional embeddings have a significant impact on the performance.

\begin{restatable}{lem}{lemthree}
\label{lem:1wl}
\textit{\name is at least as powerful as 1-WL.}
\end{restatable}
\begin{proof}
 \vskip-6pt\rev{\textit{See App.~\ref{app:proof_of_lemmathree}.}}
\end{proof}
\begin{cor}
\vskip-6pt\name is strictly more expressive than DCRNN, STGCN, \rev{and STGODE.}
\end{cor}
\renewcommand{\qedsymbol}{$\square$}
\begin{proof} \vskip-6pt This follows naturally by combining Lemmas~\ref{lem:position} and \ref{lem:1wl} since \name retains the 1-WL expressivity of DCRNN, STGCN, \rev{and STGODE}, while also being capable of distinguishing between isomorphic topologies through positional embeddings.
\end{proof}

We note that while positional embeddings have been used in the context of \gnns~\cite{pgnn,graphreach}, they do not provide 1-WL expressivity since messages are exchanged only among anchor nodes. The methodology proposed in this work is therefore unique.
    \begin{table*}[t]
    \centering
    \caption{(a) Data statistics and (b) model size.\label{tbl:dataset and modesl}}
    \vspace{-0.10in}
    \begin{subtable}[t]{.7\linewidth}
    \centering
    \caption{Details about the datasets}\label{tbl:dataset_details}
    {\footnotesize
    \begin{tabular}{lrrR{1.4cm}R{0.5cm}R{0.6cm}R{1.2cm}L{1.2cm}}
        \toprule
        \textbf{Dataset} &\textbf{ \#Nodes} & \textbf{\#Edges} & \textbf{Geographical Area (km$^2$)} & \textbf{Mean} & \textbf{Std} &\textbf{Default Seen Percent} & \textbf{\#Trips}\\
        \midrule
        Beijing~\cite{beijing} & 28465 & 64478 &  16,411 & 3.22 & 9.09 & 5\%&785709\\
        Chengdu~\cite{chengdu_data} & 3193 & 7269 &  14,378 & 4.54 & 7.19&50\%&1448940\\
        Harbin~\cite{harbin} & 6235 & 15205 &  53,068 & 77.85 & 124.11&30\%&659141\\
        \bottomrule
    \end{tabular}}
    \end{subtable}
    \begin{subtable}[t]{0.2\linewidth}
    \centering
    \caption{Number of parameters}
    \label{tab:params}
    {\scriptsize
    \begin{tabular}{lr}
        \toprule
        \textbf{Model} & \textbf{\#Parameters}\\
        \midrule
        \name & 198985 \\
        DCRNN & 372353 \\
        STNN & 313906 \\
        \rev{GraphWavent} & \rev{247660}\\
        \rev{STGODE} & \rev{558582}\\
        \rev{LocaleGN} & \rev{333636}\\
        \bottomrule
    \end{tabular}}
    \end{subtable}
\end{table*}

\noindent

\section{Experiments}
\label{sec:experiments}
In this section, we benchmark \name and establish:
\begin{itemize}
\item \textbf{Prediction Accuracy:} \name is superior compared to baselines for the task of spatio-temporal forecasting on road networks. 
\looseness=-1
\item \textbf{Ablation study:} Through extensive ablation studies, we illuminate the significance of each component of \name and provide insights into the critical role they play in the overall performance.
\item \textbf{Robustness:} We test the limits of our model to frugality in sensing, graph structure modifications at inference time and resilience to non-uniform temporal granularity across snapshots. 
\end{itemize}
The codebase of \name and datasets are available at \url{https://github.com/idea-iitd/Frigate}.


\vspace{-0.05in}
\subsection{Datasets}
We use three real-world datasets collected by tracking the GPS trajectory of taxis.
As summarized in Table~\ref{tbl:dataset_details}, the three datasets correspond to the cities of Beijing, Chengdu and Harbin~\cite{neuromlr}. We map-match~\cite{grasshopper} the raw trajectories to their corresponding road networks obtained from Openstreetmap~\cite{osm}. The traffic is aggregated into buckets of $5$ minutes and the sensor value corresponds to the number of vehicles passing through each node in a 5-minute interval. From the entire node-set, we select a subset of nodes uniformly at random as the ``seen'' nodes with installed sensors. Inference and loss calculation is performed only on the unseen nodes. The default percentage of nodes we retain as ``seen'' is shown in Table~\ref{tbl:dataset_details}. 
\looseness=-1 
\begin{figure*}
    \centering
    \subfloat[Chengdu 50\%]{\raisebox{10pt}{\includegraphics[width=.3\textwidth]{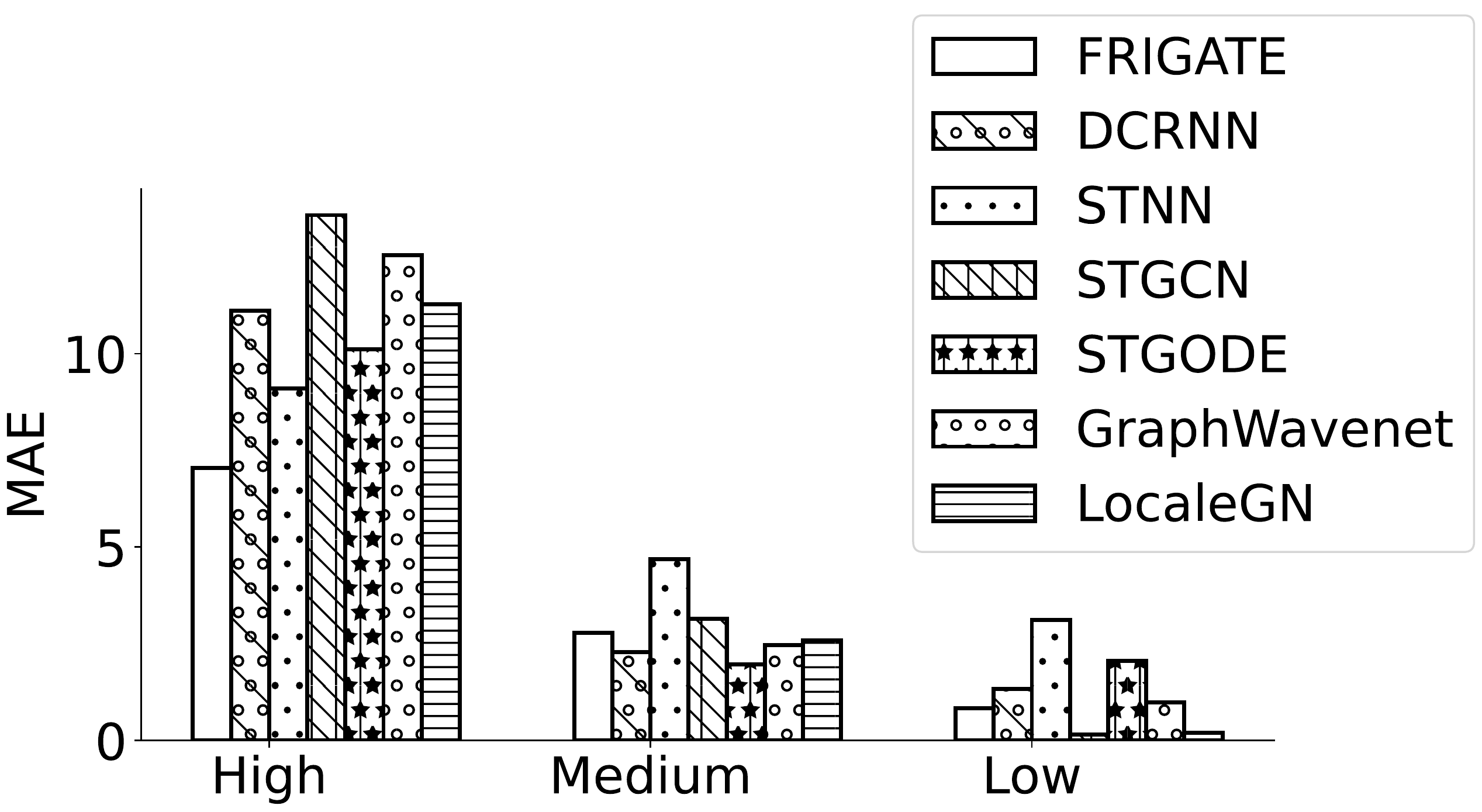}}}
    \subfloat[Harbin 30\%]{\includegraphics[width=.3\linewidth]{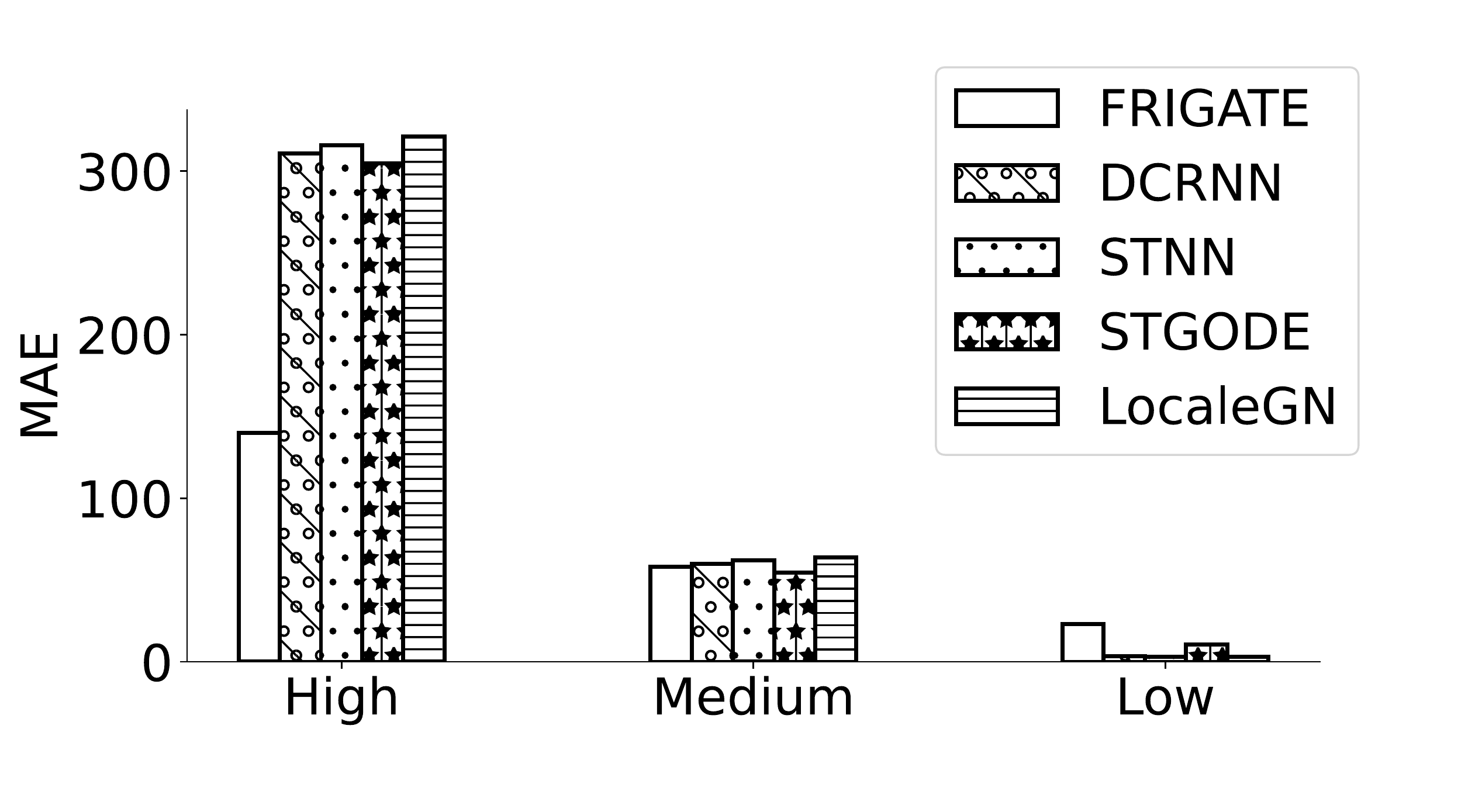}}
    \subfloat[Beijing 5\%]{\includegraphics[width=.3\linewidth]{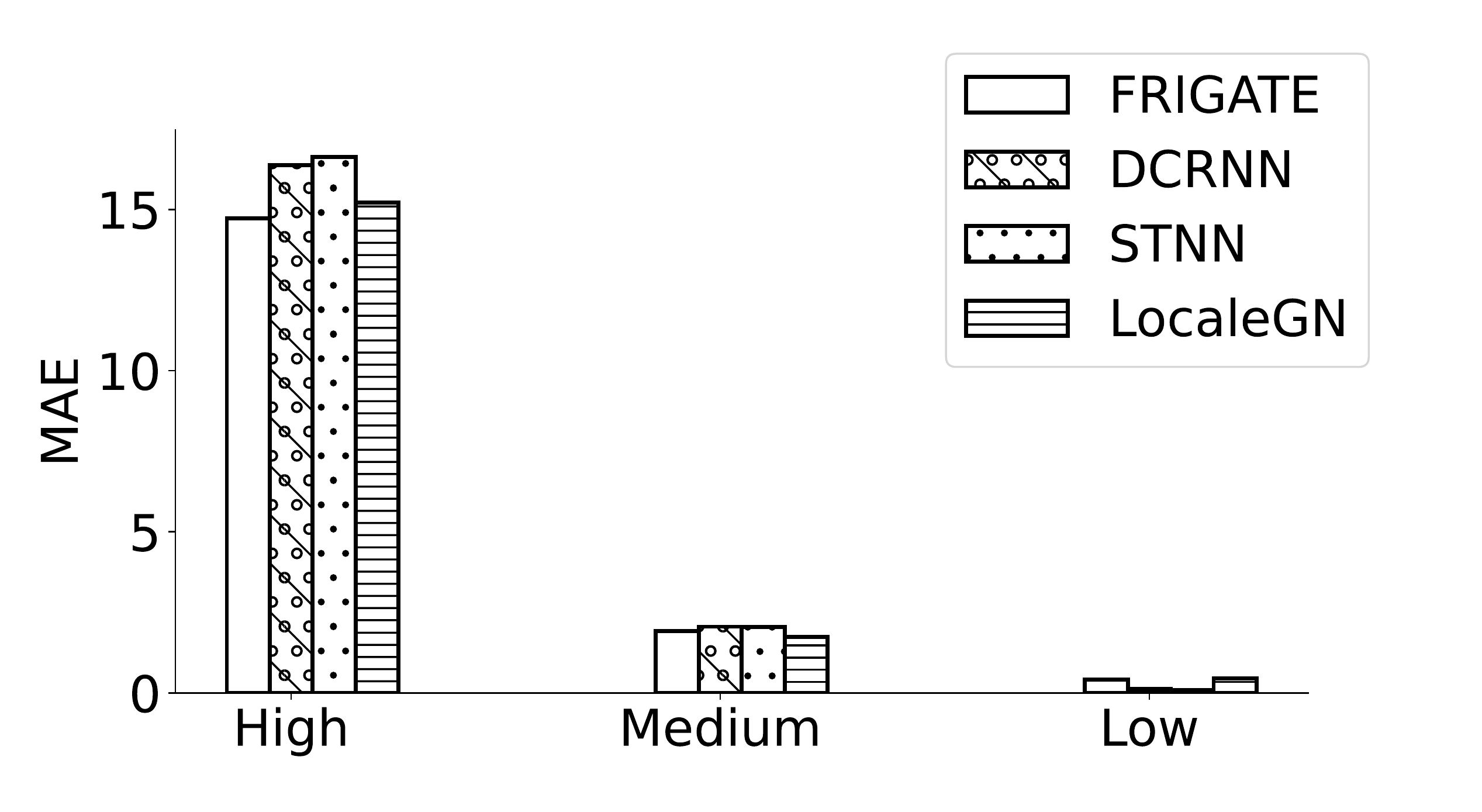}}
    \vspace{-0.15in}
    \caption{\label{fig:freq_distrib}Comparison of the distribution of error over nodes grouped by frequency of cars moving through them.}
    \vspace{-0.10in}
\end{figure*}

\vspace{-0.05in}
\subsection{Experimental Setup} 

\noindent
{\bf $\bullet$ Computational engine:} We use a system running on Intel Xeon 6248 processor with 96 cores and 1 NVIDIA A100 GPU with 40GB memory for our experiments. 

\noindent
{\bf $\bullet$ Forecasting horizon:}  In all the experiments, we choose to predict one hour of traffic information from one hour of historical traffic information. Concretely, since in the datasets the duration between two snapshots is 5 minutes, we set $\Delta=12$. Later, we also simulate what happens if the duration between snapshots is irregular, but even then we have past one hour of data and make predictions of traffic condition up to one hour in the future.

\noindent
{\bf $\bullet$ Evaluation metric:} We use mean absolute error (MAE) as the primary metric across all methods (lower is better). We also use sMAPE and RMSE for our key results in Table~\ref{tbl:giant}. In addition, we 
report a $95\%$ confidence interval around the mean by using bootstrapping on the distribution of the metric being used on the ``unseen'' nodes.%

\noindent\textbf{ $\bullet$ Training setting:} We use a $70$\%-$20$\%-$10$\% train-val-test split for training. This split is on the time dimension. So, to be explicit, during training $70\%$ of the total time series on seen nodes are used. Validation and test are both on different parts of the data both from node perspective
and timestep perspective. We stop the training of the model if it does not improve the validation loss for more than 15 epochs. Further, we have  varied percentage of training nodes in various experiments to check the fidelity of the FRIGATE, as indicated in the results.
\looseness=-1

\noindent
{\bf $\bullet$ Baselines:} \rev{We consider DCRNN~\cite{dcrnn}, STGCN~\cite{stgcn}, LocaleGN~\cite{li2022few} and STNN~\cite{stnn} as our baselines.} \rev{We adapt GraphWavenet~\cite{wavenet} and STGODE~\cite{stgode} to our setting by removing node embedding matrices or Dynamic Time Warping based graph computation so that they support the triple objectives outlined in Table~\ref{tbl:baselines}.} For all \rev{six} baselines, we obtain the official code-base released by the authors.

\noindent
{\bf $\bullet$ Parameter settings:} We use $16$ anchor nodes to calculate Lipschitz embeddings for all graphs. We use $10$ layers of \gnn in \name. 

\vspace{-0.10in}
\subsection{Inference Accuracy}
 In Table~\ref{tbl:giant}, we present the MAE obtained by \name and all other baselines on all three datasets. We observe that \name  consistently outperforms all baselines. On average, the MAE in \name is more than 25\% lower than the closest baseline. Among the baselines, \rev{LocaleGN and STGODE perform the best, followed by DCRNN, followed by GraphWavenet}, followed by STNN and then STGCN. We further note that STGCN fails to train on Harbin and Beijing even after 48 hours (the largest dataset evaluated in STGCN~\cite{stgcn} was of 1026 nodes and it only considered weekday traffic). \rev{Likewise, GraphWavenet also fails to train on Harbin and Beijing (the largest dataset evaluated in GraphWavenet~\cite{wavenet} was of 325 nodes).} Now, to better contextualize the results, let us compare the MAE with the standard deviation of sensor values reported in Table~\ref{tbl:dataset_details}. We observe a clear correlation of MAE with the standard deviation, which explains why all techniques perform comparatively poorer in Harbin. Due to the substantially inferior performance of STGCN and its inability to scale on large datasets, subsequent experiments only consider DCRNN, STNN, \rev{STGODE, and LocaleGN} as baselines.

To further diagnose the performance pattern among the benchmarked techniques, we segregate the nodes into three buckets based on the number of cars going through them and plot the distribution of errors within these three buckets. Fig.~\ref{fig:freq_distrib} presents the results. Here, ``High'', ``Medium'' and ``Low'' corresponds to the top-33 percentile, 33-66 percentile, and bottom-33 percentile, respectively. We derive three key insights from this experiments. First, across all techniques, the MAE reduces as we move towards nodes handling lower traffic. This is natural, since these nodes have low variation in terms of traffic volume. In contrast, high-traffic nodes undergo more fluctuation depending on the time of the day, weekday vs. weekends, etc. Second, we note that \name performs better than all other models on high frequency nodes, good on medium frequency nodes, and worse than the rest in low frequency nodes. Overall, it may be argued that doing better in high and medium-frequency nodes are more important since they handle a large majority of the traffic, where \name mostly outperforms other techniques. Finally, one of the key reasons of under-performance in DCRNN, STNN, \rev{and STGODE} is that they over-smooth the high-frequency nodes into inferring lower values as the majority of the nodes are of low-frequency (recall power-law distribution from Fig.~\ref{fig:freq_distrib}). Positional embeddings and deep layering through gating enables \name to avoid over-smoothing. As we established in Section~\ref{sec:characterization}, these design choices make \name provable more expressive. We further investigate the impact of these two components in our ablation study in Section~\ref{sec:ablation}. 
\begin{table}[b!]
        \vspace{-0.20in}
        \centering
        \caption{\label{tbl:giant}Performance comparison of different approaches for traffic volume prediction. OOT stands for Out of Time \rev{and OOM stands for Out of Memory}. The lowest MAE (best accuracy)\rev{, lowest sMAPE and lowest RMSE are} highlighted in \textbf{bold} for each dataset.}
        {\footnotesize
        \begin{tabular}{llrrr}
        \toprule
        \textbf{Model} & \textbf{Dataset} & \textbf{MAE} & \rev{\textbf{sMAPE}} & \rev{\textbf{RMSE}}\\
        \midrule
        \name & Chengdu  & $\mathbf{3.547\pm 0.20}$ & $\rev{\mathbf{131.17\pm 3.39}}$ & $\rev{\mathbf{5.93\pm0.31}}$\\
        STNN &   & $5.633\pm 0.18$ & $\rev{199.86\pm0.00}$ & $\rev{9.50\pm0.50}$\\
        DCRNN &   & $4.893\pm 0.26$& $\rev{138.17\pm2.07}$ & $\rev{8.10\pm0.48}$\\
        \rev{LocaleGN} & & $\rev{4.597\pm0.22}$ & $\rev{192.80\pm0.19}$ & $\rev{8.57\pm0.33}$\\
        \rev{STGODE} & & $\rev{4.693\pm0.19}$ & $\rev{147.12\pm2.25}$ & $\rev{7.72\pm0.39}$\\
        STGCN &   & $5.712\pm 0.33$& $\rev{198.38\pm0.00}$ & $\rev{9.47\pm0.50}$\\
        \rev{GraphWavenet} & & $\rev{5.308\pm0.30}$ & $\rev{147.77\pm1.81}$ & $\rev{8.89\pm0.43}$\\
        \midrule
        \name & Harbin  & $\mathbf{73.559\pm 2.04}$ & $\rev{\mathbf{93.00\pm1.89}}$ & $\rev{\mathbf{105.01\pm3.51}}$\\
        STNN &   & $126.305\pm 4.54$ & $\rev{199.68\pm0.02}$ & $\rev{206.91\pm6.93}$\\
        DCRNN &   & $124.109\pm 5.06$ & $\rev{185.73\pm0.68}$ & $\rev{204.75\pm6.36}$\\
        \rev{LocaleGN} & & $\rev{128.674\pm5.10}$ & $\rev{169.54\pm1.43}$ & $\rev{207.36\pm6.58}$\\
        \rev{STGODE} & & $\rev{122.493\pm4.97}$ & $\rev{152.28\pm1.27}$ & $\rev{198.81\pm6.74}$\\
        STGCN &   & OOT & \rev{OOT} & \rev{OOT}\\
        \rev{GraphWavenet} & & \rev{OOM} & \rev{OOM} & \rev{OOM}\\
        \midrule
        \name & Beijing  & $\mathbf{5.651\pm 0.11}$ & $\rev{\mathbf{169.73\pm0.45}}$ & $\rev{\mathbf{11.87\pm0.30}}$\\
        STNN &   & $6.215\pm 0.12$ & $\rev{199.99\pm0.01}$ & $\rev{12.82\pm0.27}$\\
        DCRNN &   & $6.122\pm 0.14$ & $\rev{195.56\pm0.15}$ & $\rev{12.63\pm0.24}$\\
        \rev{LocaleGN} & & $\rev{5.759\pm0.13}$ & $\rev{172.73\pm0.33}$ & $\rev{12.10\pm0.03}$\\
        \rev{STGODE} & & \rev{OOM} & \rev{OOM} & \rev{OOM}\\
        STGCN &   & OOT & \rev{OOT} & \rev{OOT}\\
        \rev{GraphWavenet} & & \rev{OOM} & \rev{OOM} & \rev{OOM}\\
        \bottomrule
        \end{tabular}}
        \vspace{-0.10in}
\end{table}
%
\begin{figure}[b]
\vspace{-0.10in}
    \centering
    \subfloat[Chengdu]{
    \hspace*{0.25cm}\includegraphics[width=1\linewidth]{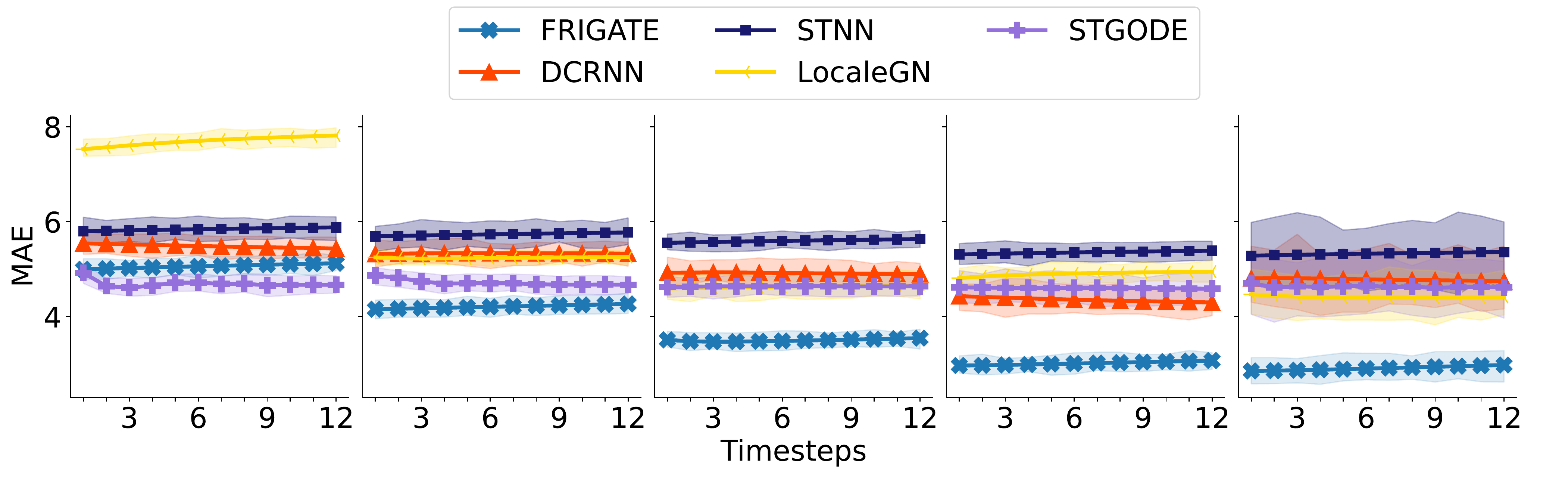}}\\
    \subfloat[Harbin]{
    \includegraphics[width=1\linewidth]{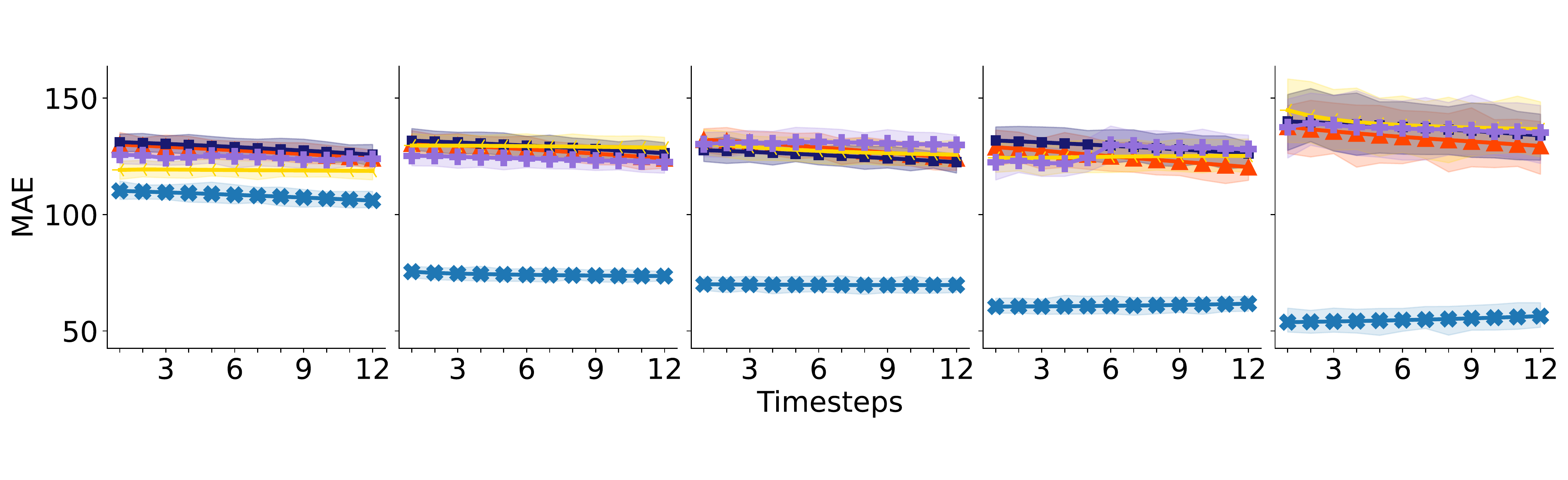}}
    \vspace{-0.10in}\caption{\label{fig:pct_variation}From left to right, we progressively increase the volume of seen nodes and measure MAE in (a) Chengdu and (b) Harbin. The envelopes signify the 95\% confidence bound of the MAEs.}
\end{figure}
\begin{figure*}
    \centering
    \subfloat[]{
    \includegraphics[width=0.3\linewidth]{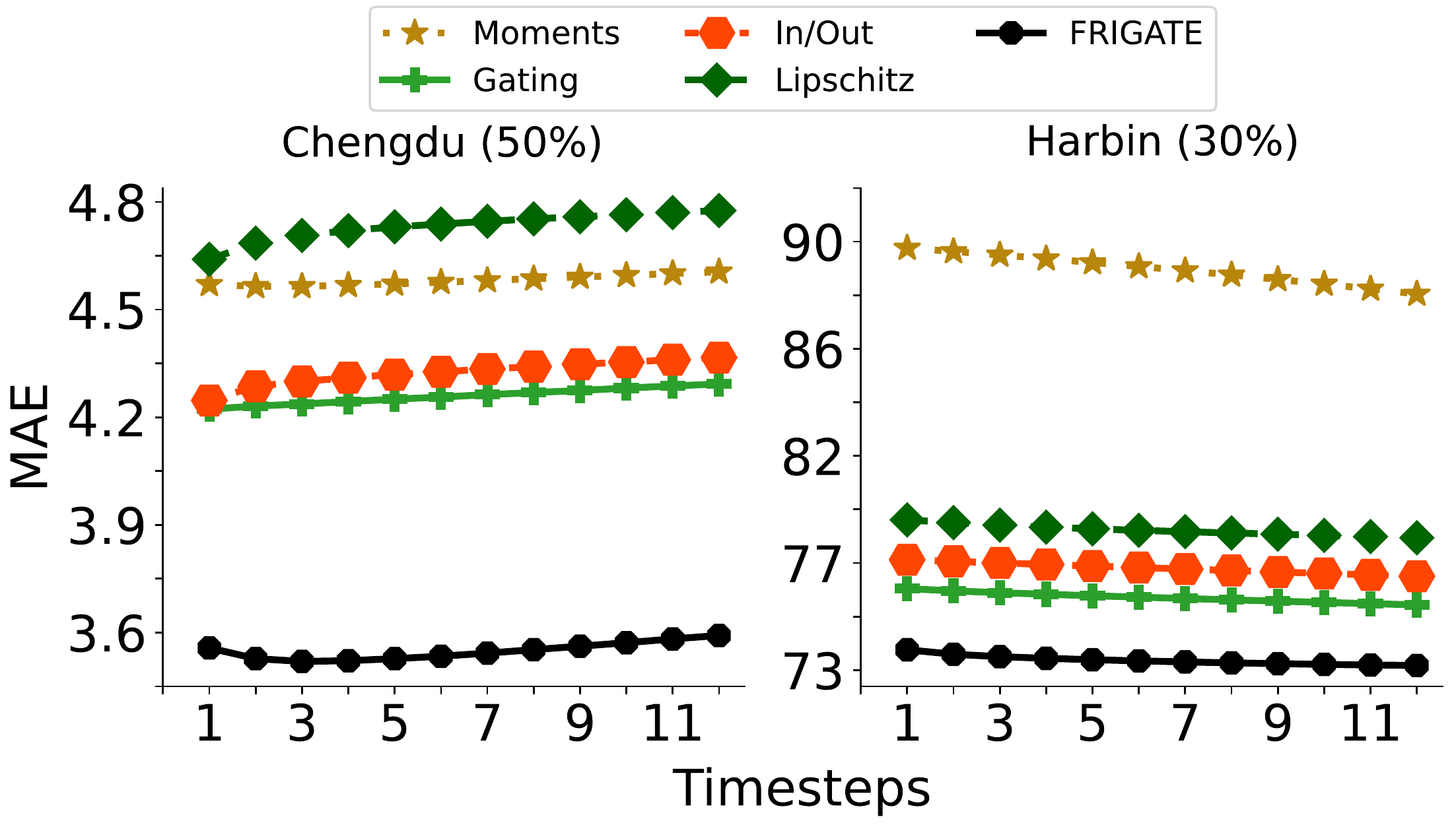}
    \label{fig:ablation}
    }
    \subfloat[]{
     \includegraphics[width=0.3\linewidth]{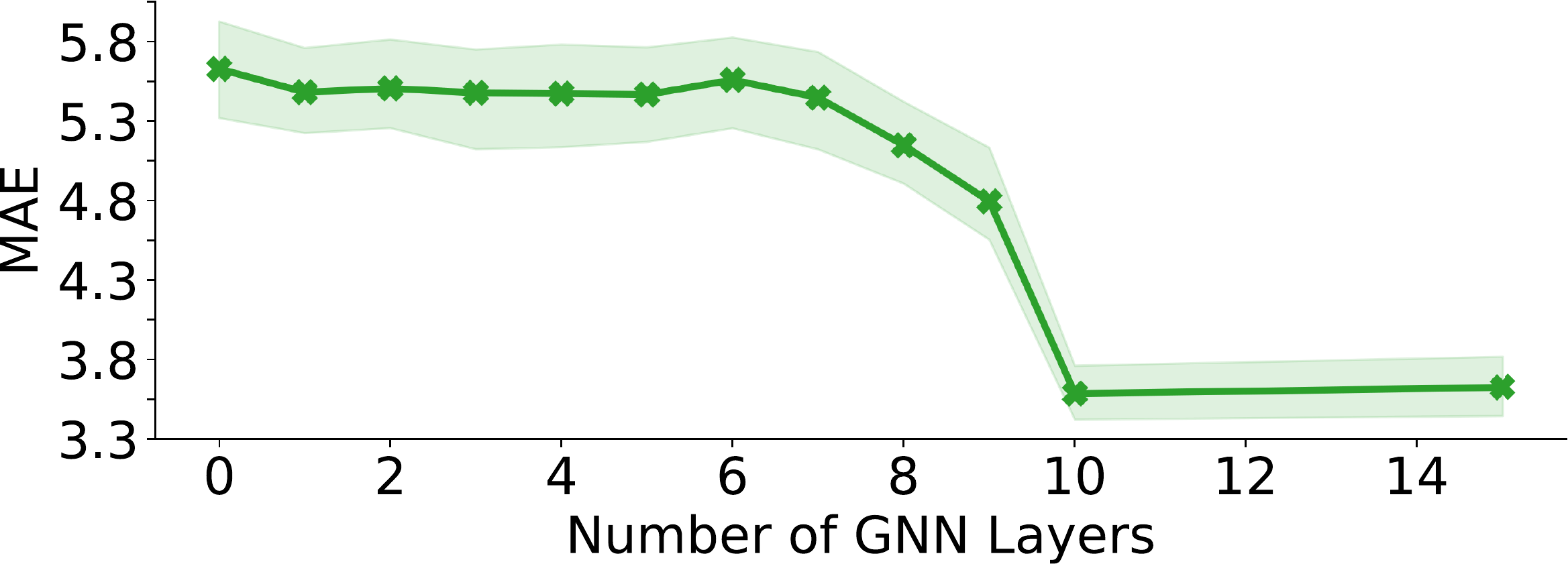}
    \label{fig:layer_variations}
    }
    \subfloat[]{
    \includegraphics[width=0.3\linewidth]{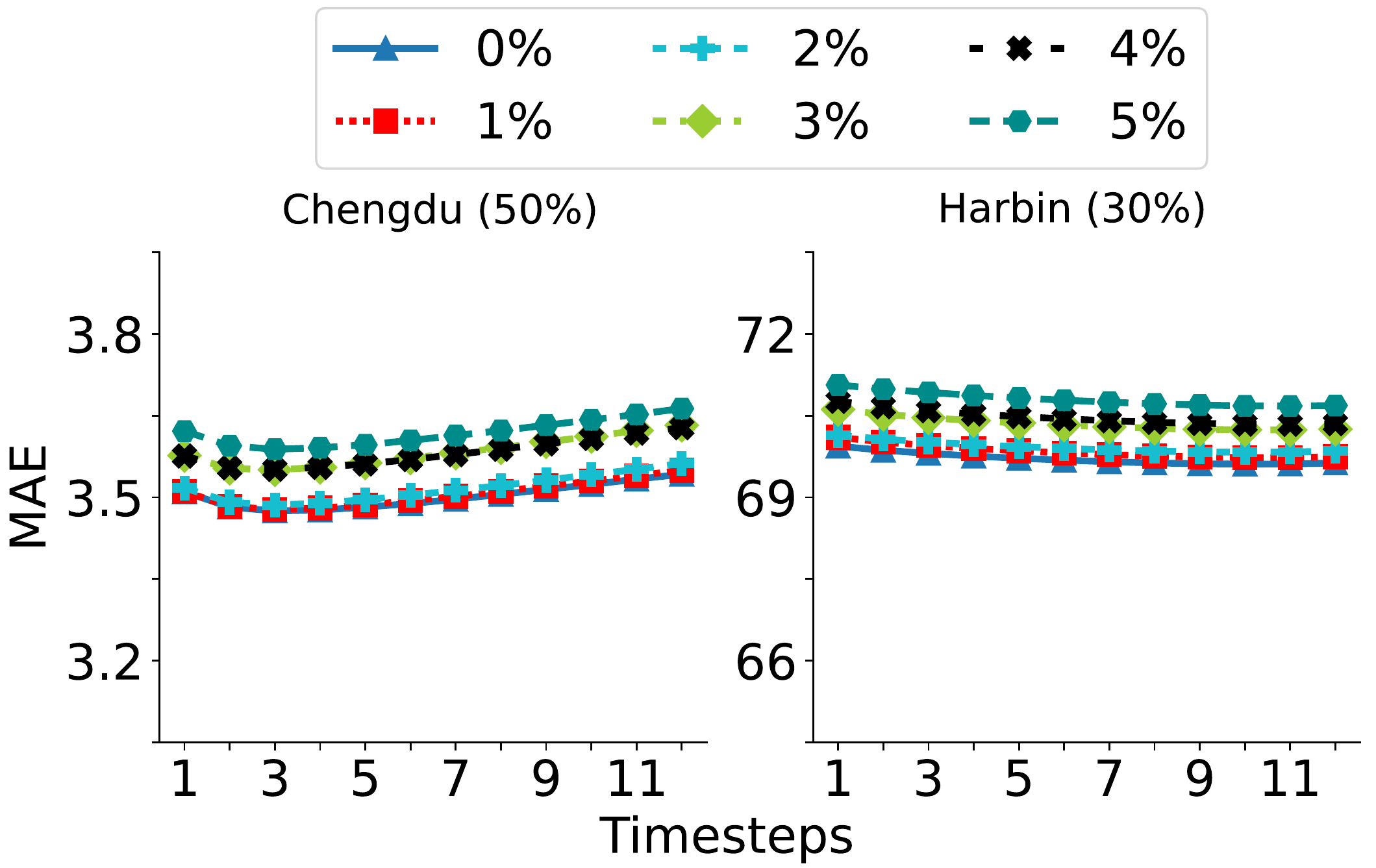}
    \label{fig:gchanges}
    }
    \vspace{-0.10in}
    \caption{ (a) Ablation study. (b) Impact of \gnn layer variations on Chengdu 50\% (c) Model resilience to inference time changes on road network topology.}
\end{figure*}
\vspace{-0.10in}
\subsection{Impact of Volume of ``Seen'' Nodes}
 As in any machine learning task, we expect the accuracy to improve with larger volume of seen nodes. This increases training data, as well as induces a higher likelihood of an unseen node being close to a seen node. 
Towards that objective, we vary the number of nodes that are ``seen'' by the models from $10\%$ to $90\%$ of the total
number of nodes in the respective road networks and measure MAE against forecasting horizon. Fig.~\ref{fig:pct_variation} presents the results. 
We observe that \name performs significantly better because of more informative
priors and inherent inductivity. Furthermore, \rev{the baselines} need considerably more data to reach the same performance. For Chengdu, DCRNN trained on 50\% of the graph beats our model trained on only 10\% of the graph while on Harbin, \rev{none of} the models ever even surpass our model trained at only 10\% of the graph. And as the data increases, the gap between our model and the baselines increases. This economic use of data without losing the ability to generalize is a desirable property that \name contains. 

Another interesting trend we note is that the performance slightly deteriorates in \rev{the baselines} from 70\% to 90\%. It is hard to pinpoint the exact reason. We hypothesize two factors as possible reasons. First, with higher volumes of training data, there is a stronger chance of over-fitting. In addition, we note that the confidence interval expands as the percentage of seen nodes increases since the sample size of test nodes decreases. This trend, which is consistent with statistical theory, means that at low volumes of test sets, the results have higher variability. \name does not suffer from this trend, which means it does not overfit. It is well-known in machine learning theory, that tendency to overfit is correlated to the number of parameters. In this regard, we draw attention to Table~\ref{tab:params}, which shows that \name has almost 50\% and 33\% smaller parameter set than DCRNN and STNN, respectively. Also, \name utilizes moments as a robust inductive bias, reducing overfitting risks. 

%
%
\subsection{Ablation Study}
\label{sec:ablation}
In our ablation study, we systematically turn off various components of \name and measure the impact on MAE. Fig.~\ref{fig:ablation} presents the results on Chengdu and Harbin.

\noindent\textbf{Gating:} 
Here, we replace the gated convolution with a GraphSAGE layer~\cite{graphsage}, leaving rest of the architecture intact. We see a clear
increase in MAE indicating the importance of gated convolution layer.
\looseness=-1

\noindent\textbf{Positional embeddings:} 
To understand its utility on performance, we remove Lipschitz embeddings as features of the node, and thus remove it as an input to the calculation of gating. We see a significant drop in performance, with Chengdu being more pronounced where Lipschitz has the highest drop. This result empirically demonstrates the value of positional embeddings in time-series forecasting on road networks. 

\noindent\textbf{In-Out aggregation:} In \name, we separately aggregate messages from incoming and outgoing neighbors. Here, we measure the impact of aggregating into a single vector. We observe an increase in MAE, showcasing the need for direction-based aggregation.
\looseness=-1

\noindent\textbf{Moments:} We remove the moments from the model and keep everything else in the architecture the same. We see a big drop in performance on both datasets, with Harbin being more pronounced. 
This establishes the importance of having an informative prior.

\noindent\textbf{Number of GNN layers:} To understand the effect of \gnn layers on performance of \name, we vary this parameter while training on Chengdu 50\% dataset. As visible in Fig.~\ref{fig:layer_variations}, the performance saturates at $10$, which we use as our default value. 
\looseness=-1


%
%
\vspace{-0.10in}
\subsection{Robustness and Resilience}
In this section, we analyze the robustness and resilience of \name to changes to network topology and irregular topological sensing. In these experiment, we train \name on the original datasets, and then slightly perturb the topology or temporal granularity. We evaluate inference performance on this perturbed dataset. Note that we do not re-train the model after perturbations.

\noindent
\textbf{Topology change: } To simulate the real-world situation where the road network might change due to road blocks or opening of new roads, we
first select the volume of perturbations to be introduced. Assuming this to be $X\%$, we change the network by randomly dropping a maximum of $\frac{X}{2}$\% of the edges. In addition, we create an equal number edges among unconnected nodes whose distance is within a threshold. The distances for these edges are sampled, with replacement, from the original distribution of edge distance. We then vary $X$ and measure the impact on MAE. 
As visible in Fig.~\ref{fig:gchanges}, \name is resilient to changes to the road network with minimal drop in accuracy.
\looseness=-1

%
\begin{table}[t]
    \centering
    \caption{\label{tbl:time_granularity}Impact of irregular temporal granularity on MAE.}
    {
    \begin{tabular}{llc}
        \toprule
        Time series & Dataset & MAE\\
        \midrule
        Regular (original) & Chengdu (50\%) & $3.547\pm 0.23$\\
        Irregular (perturbed) & Chengdu (50\%) & $3.582\pm 0.17$\\
        \midrule
        Regular (original) & Harbin (50\%) & $69.700\pm 3.21$\\
        Irregular (perturbed) & Harbin (50\%) & $69.834\pm 2.78$\\
        \bottomrule
    \end{tabular}}
    \vspace{-0.20in}
\end{table}
\noindent
{\bf Irregular temporal sensing: }In this experiment, we try to break our model by changing the time step granularity at test time. We randomly drop 33.33\% of snapshots in $\rns_{[t-\Delta,t]}$, effectively reducing the time sequence length and also changing the granularity at which the data is captured. The results are tabulated in table~\ref{tbl:time_granularity}, where we observe negligible increase in MAE. The results indicate that the model is resilient to changes in the granularity of snapshots.
\section{Conclusion}
In this paper, we have proposed a new spatio-temporal \gnn, called \name, to forecast network-constrained time-series processes on road networks. We show through experiments on real-world traffic datasets that \name significantly outperforms existing baselines. In addition, \name is robust to several practical challenges such as partial sensing across nodes, absorb minor updates to road network topology and resilience to irregular  temporal sensing.
 The core competency of \name originates from its novel design that incorporates Lipschitz embedding to encode position, sigmoid gating to learn message importance and enable pathways for long-range communication across nodes, directional aggregation of messages, and strong priors in the form of moments of the time-series distribution. In addition, \name is built from the ground-up keeping inductivity as a core objective. On the whole, \name takes us closer to deployable technology for practical workloads. 

 \noindent
\section*{Acknowledgments} This research was supported by Yardi School of AI, IIT Delhi. Sayan Ranu acknowledges the Nick McKeown Chair position endowment. 
\noindent

\bibliographystyle{ACM-Reference-Format}
\FloatBarrier
\clearpage
\balance
\bibliography{ref}
\clearpage
\appendix
\counterwithin{figure}{section}
\counterwithin{table}{section}
\section{Notations}
\begin{table}[h]
    \centering     
    \caption{Notations used in the paper\label{tab:notation}}
        \begin{tabular}{m{0.25\linewidth}m{0.6\linewidth}}                    
        \toprule                                                               
        \textbf{Symbol} & {\textbf{Meaning}} \\ 
        \midrule                                                              
         $\rns$ & The road network stream\\
         \cmidrule(lr){1-2}                                                    
         $\CG_t$& Road network snapshot at time $t$ \\          
        \cmidrule(lr){1-2}
        $\rns_{[i,j]},\:i\le j$ & $\{\CG_i,\CG_{i+1},\dotsc,\CG_j\}\subseteq\rns$\\
        \cmidrule(lr){1-2}  
        $\CV_t$ & Vertex set at timestep $t$\\
        \cmidrule(lr){1-2}
        $\CV^{tr}$ & Set of vertices to be trained on\\
        \cmidrule(lr){1-2}
         $\CE_t$ & Edge set at timestep $t$\\         
        \cmidrule(lr){1-2}
        $\delta$ & \textit{Haversine} distance function\\
        \cmidrule(lr){1-2}
        $\tau_t$ & Sensor readings $\forall$ nodes at time $t$\\
        \cmidrule(lr){1-2}
        $T$ & Number of snapshots of the graph\\
        \cmidrule(lr){1-2}
        $\Delta$ & Prediction horizon: how many timesteps in future to predict\\
        \cmidrule(lr){1-2}
        $\Psi_\Theta$ & The model with parameters $\Theta$\\
        \cmidrule(lr){1-2}
        $\mathcal{P}$ & Permutation function\\
        \cmidrule(lr){1-2}
         $\mathcal{N}(v)$ & Set of nodes in the neighborhood of node $v$\\
         \cmidrule(lr){1-2}
         $\CL_v$&Positional embedding of node $v$\\
         \cmidrule(lr){1-2}
         $\CL_{v_i,v_j}$ & Positional embedding of edge $(v_i,v_j)$\\
         \cmidrule(lr){1-2}
         $\h_v^\ell$&Intermediate representation of node $v$ at layer $\ell$ in \gnn\\
         \cmidrule(lr){1-2}
         $\z_v^k$ & Final representation of node $v$ in $k^{th}$ stack output by \gnn\\
         \cmidrule(lr){1-2}
         $\mbd{s}_v^k$ & Hidden state + cell state of $\oplstm_\text{Enc}$ at step $k$\\
         \cmidrule(lr){1-2}
         $\mbd{m}_v^t$ & Representation of moments of $\CN(v)$ at time $t$\\
         \cmidrule(lr){1-2}
         $\mbd{q}_v^k$ & Hidden state + cell state of $\oplstm_\text{Dec}$\ at step $k$\\
         \cmidrule(lr){1-2}
         $\hat{y}_v^k$ & Prediction at $k^{th}$ step from $\oplstm_{\text{Dec}}$\\
         \cmidrule(lr){1-2}
         $\cdot\Vert\cdot$ & Concatenation operator\\
         \cmidrule(lr){1-2}
         $sp(u,v)$ & Shortest path between $u$ and $v$\\
         \cmidrule(lr){1-2}
         $\lVert\cdot\rVert_2$ & L$_2$ norm\\
        \bottomrule
        \end{tabular}
\end{table}
\section{Inference time}
\FloatBarrier
The inference times of the three models are shown in Table~\ref{tab:inference}. \name is faster than STNN but slower than DCRNN. However, we note that even though \name is slower than DCRNN it is still fast enough to be deployed for real-world workloads.
\begin{table}[!htbp]
    \centering
    \caption{Inference times}
    \label{tab:inference}
    \begin{tabular}{lr}
        \toprule
        Model & Time(s) \\
        \midrule
         DCRNN &  0.0081\\
         STGODE & 0.0164\\
         \name & 0.0207\\
         STNN & 0.0375\\
         LocaleGN & 0.0492\\
         \bottomrule
    \end{tabular}
\end{table}
\section{Parameter size}
\begin{table}[H]
    \centering
    \caption{Dimensionality of the parameters used in \name.}
    \label{tab:parameters}
    \begin{tabular}{cc}
        \toprule
        \textbf{Parameter} & \textbf{Shape}\\
        \midrule
        $\mbd{W}_1^\ell$&$\mathbb{R}^{3d\times d}$\\
        $\mbd{w}_\delta^T$&$\mathbb{R}^{d_\delta}$\\
        $\mbd{W}_\CL^\ell$&$\mathbb{R}^{2d_L+d\delta\times d_{L_e}}$\\
        $\mbd{w}^\ell$&$\mathbb{R}^{d_{L_e}}$\\
        $\mbd{w}^T_\tau$&$\mathbb{R}^{d_\tau}$\\
        $\mlp_1$&$\mathbb{R}^{d_{Lenc}}$\\
        $\mlp_2$&$\mathbb{R}^{d_{dec}}$\\
        $\mlp_3$&$\mathbb{R}^{d_{moments}}$\\
        $\mlp_4$&$\mathbb{R}^{d_{dec}+d_{moments}}$\\
        \bottomrule
    \end{tabular}
\end{table}
\section{Other results}
\begin{figure}[!htbp]
    \centering
    \includegraphics[width=\linewidth]{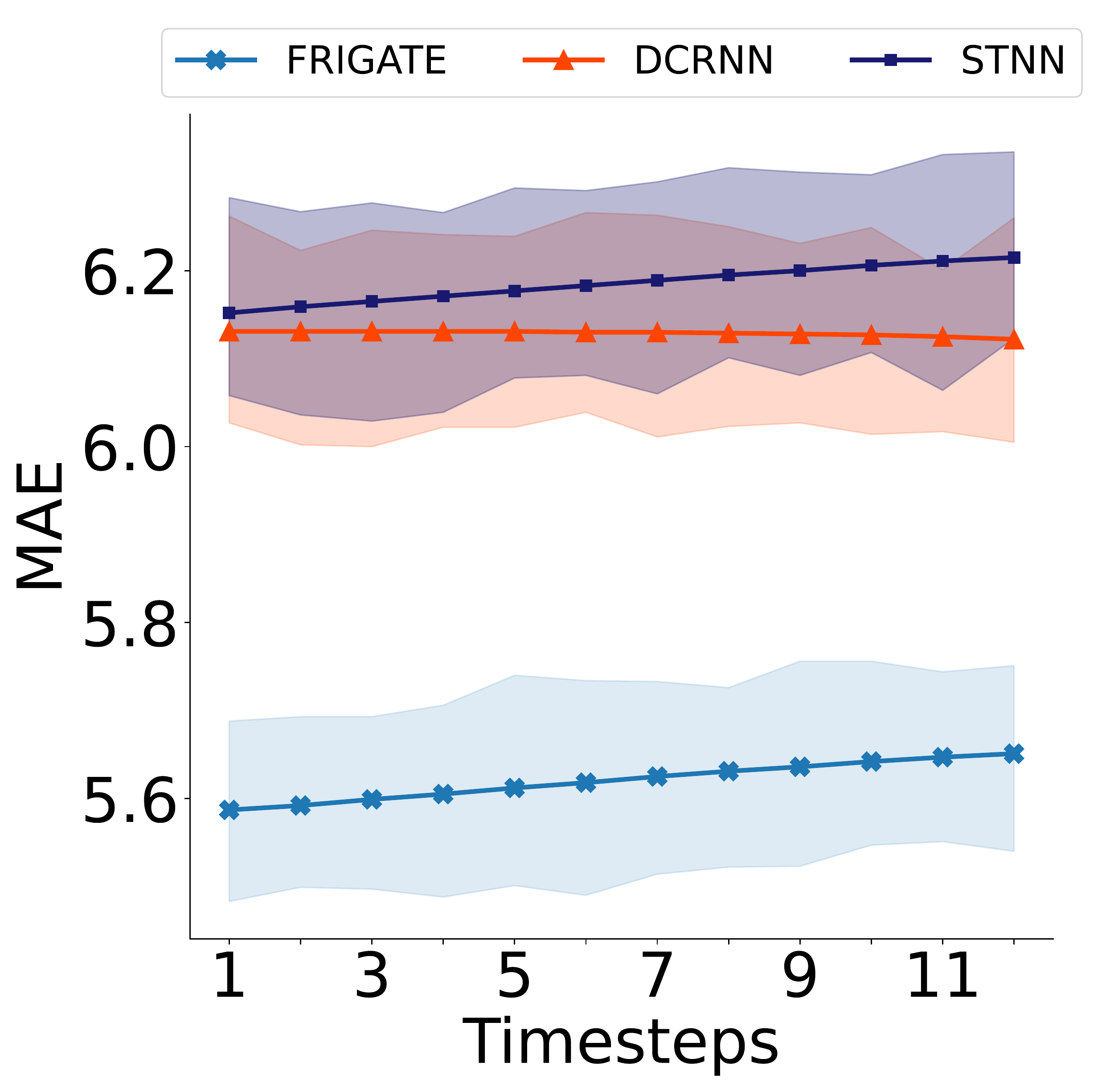}
    \caption{Prediction on Beijing (5\%) dataset}
    \label{fig:beijing_prediction}
\end{figure}
Fig.~\ref{fig:beijing_prediction} shows the MAE of the three models on Beijing (5\%) dataset against the prediction horizon.
\renewcommand{\qedsymbol}{$\square$}
\section{Proof of Lemma~\ref{lem:metric_dist}}
\label{app:proof_lem_one}
\lemone*
\begin{proof}
We need to show \textbf{(1)} Symmetry: $d_X(u,v)=d_X(v,u)$, \textbf{(2)} Non-negativity: $d_X(u,v)\geq 0$, \textbf{(3)} $d_X(u,v)=0$ iff $u=v$ and \textbf{(4)} Triangle inequality: $d_X(u,v)\leq d_X(u,w)+d_X(w,v)$. 
 
 We omit the proofs of first three properties since they are trivial. We use proof-by-contradiction to establish triangle inequality. 
 \vspace{-0.05in}
 \begin{alignat}{2}
 \label{eq:contradiction}
 \text{Let us assume }d_X(u,v)&> d_X(u,w)+d_X(w,v)\\
 \nonumber
 or, sp(u,v)+sp(v,u)&>sp(u,w)+sp(w,u)+sp(w,v)+sp(v,w)
 \end{alignat}
\noindent
 From the definition of shortest paths, $sp(u,v)\leq sp(u,w)+sp(w,v)$ and $sp(v,u)\leq sp(v,w)+sp(w,u)$. Hence, Eq.~\ref{eq:contradiction} is a contradiction.
\end{proof}

\section{Proof of Lemma~\ref{lem:1wl}}
\label{app:proof_of_lemmathree}
\lemthree*
\begin{proof}
 \gin~\cite{gin} is as powerful as 1-WL~\cite{gin}. This power is induced by the \textsc{Sum-Pool} aggregation since sum-pool is \textit{injective} function, i.e., two separate aggregation over messages would be identical if and only if the input messages are identical~\footnote{ As in \gin~\cite{gin}, we assume countable features on nodes.}. \name can also model sum-pool, and in the more general case, an injective message aggregation, whenever the sigmoid gate over all edges is some value $x\neq 0$ (Recall Eq.~\ref{eq:sigmoid}). This happens if $\w^{\ell}$ in Eq.~\ref{eq:edgescalar} is a zero vector, the bias $b$ in 
 Eq.~\ref{eq:edgescalar} is non-zero and the submatrix in $\W^1_1$ of Eq.~\ref{eq:hl} applying linear transformation to Lipschitz embeddings of nodes is $0$.
\end{proof}
\end{document}